\pdfoutput=1

\documentclass[11pt]{article}

\usepackage[]{acl}

\usepackage{times}
\usepackage{latexsym}

\usepackage[T1]{fontenc}

\usepackage[utf8]{inputenc}

\newcommand{\Sref}[1]{\S\ref{#1}}
\newcommand{\method}{\textsc{Quote-Tuning}} 
\newcommand{\pipt}{\pi_\text{ref}}

\newcommand\blfootnote[1]{%
  \begin{NoHyper}%
  \renewcommand\thefootnote{}\footnote{#1}%
  \addtocounter{footnote}{-1}%
  \end{NoHyper}%
}

\newcommand{\jack}[1]{{\color{orange}[JZ: #1]}}

\newcommand{\paravs}{\vspace{-1mm}}

\usepackage{microtype}

\usepackage{inconsolata}

\usepackage{amsfonts}
\usepackage{booktabs,arydshln}
\usepackage{multirow}
\usepackage{graphicx}
\usepackage{url}
\usepackage{hyperref}
\usepackage{wrapfig}
\usepackage{amsmath}
\usepackage{bbm}
\usepackage{algorithm}
\usepackage{algorithmicx}
\usepackage[noend]{algpseudocode}
\usepackage{epigraph}
\usepackage{xcolor}
\usepackage{soul}
\usepackage{subfig}

\definecolor{quoted}{HTML}{0066ff}

\sethlcolor{quoted!10}
\DeclareRobustCommand{\quoteda}[1]{{\sethlcolor{quoted!60}\hl{#1}}}
\DeclareRobustCommand{\quotedb}[1]{{\sethlcolor{quoted!25}\hl{#1}}}
\DeclareRobustCommand{\quotedc}[1]{{\sethlcolor{quoted!10}\hl{#1}}}

\makeatletter
\def\adl@drawiv#1#2#3{%
    \hskip.5\tabcolsep
    \xleaders#3{#2.5\@tempdimb #1{1}#2.5\@tempdimb}%
            #2\z@ plus1fil minus1fil\relax
    \hskip.5\tabcolsep}
\newcommand{\cdashlinelr}[1]{%
\noalign{\vskip\aboverulesep
       \global\let\@dashdrawstore\adl@draw
       \global\let\adl@draw\adl@drawiv}
\cdashline{#1}
\noalign{\global\let\adl@draw\@dashdrawstore
       \vskip\belowrulesep}}
\makeatother

\newcommand*\Let[2]{\State #1 $\gets$ #2}
\algrenewcommand\algorithmicrequire{\textbf{Input:}}
\algrenewcommand\algorithmicensure{\textbf{Output:}}
\newcommand{\dataportrait}{\textsc{Data Portrait}}
\algnewcommand{\parState}[1]{
  \parbox[t]{\dimexpr\linewidth-\algmargin}{\strut #1\strut}}

\newcommand{\aspace}{\hspace{1.05em}}

\title{Tuning Language Models to Quote}
\title{\method: Aligning Language Models to Quote\\ from Pre-Training Data}
\title{\textit{Grounded by Design}: Aligning Language Models \\ to Quote from Pre-Training Data}
\title{\method: Improving Large Language Model Verifiabilty by Design}
\title{\method: Aligning Language Models to Quote Improves Verifiabilty by Design}
\title{\textit{Verifiable by Design}: Aligning Language Models to Quote from Pre-Training Data}
\title{\textit{Verifiable by Design}: Aligning Language Models to Quote}
\title{Encouraging Language Models to Quote from Pre-Training Data}
\title{Aligning Language Models to Quote from Pre-Training Data}
\title{\method: \\Aligning Language Models to Quote from Pre-Training Data}
\title{
\vspace*{-0.5in}
{{\small \hfill NAACL '25}\\
\vspace*{.25in}} 
\textit{Verifiable by Design}: \\Aligning Language Models to Quote from Pre-Training Data}

\author{%
    Jingyu Zhang\aspace
    Marc Marone\aspace
    Tianjian Li\aspace
    Benjamin Van Durme$^\heartsuit$\aspace
    Daniel Khashabi$^\heartsuit$\\
    Johns Hopkins University \\ 
    Baltimore, MD \\
    \texttt{\{jzhan237,mmarone1,tli104\}@jhu.edu}
}

\begin{document}
\maketitle
\begin{abstract}
To trust the fluent generations of large language models (LLMs), humans must be able to \emph{verify} their correctness against trusted external sources. Recent efforts, such as providing citations via retrieved documents or post-hoc provenance, enhance verifiability but provide no guarantees of their correctness. 
To address these limitations, we aim to improve verifiability with a different philosophy: \emph{trivializing the verification process by developing models that quote \underline{verbatim} statements from trusted sources in their pre-training data.} 

We propose \method, which demonstrates the feasibility of aligning models to quote. The core of \method\ is a fast membership inference function that efficiently verifies text against trusted corpora. We leverage this tool to design a reward function to quantify quotes in model responses and curate datasets for preference learning.
Experiments show that \method\ significantly increases verbatim quotes from high-quality documents by up to 130\% relative to base models while maintaining response quality. \method\ is applicable in different tasks, generalizes to out-of-domain data and diverse model families, and provides additional benefits to truthfulness. 
Our method not only serves as a hassle-free method to increase quoting but also opens up avenues for improving LLM trustworthiness through better verifiability.\footnote{
\href{https://github.com/JHU-CLSP/verifiable-by-design}{\texttt{github.com/JHU-CLSP/verifiable-by-design}}}
\blfootnote{$\heartsuit$ Equal advising.}

\end{abstract}

\setlength\epigraphwidth{.75\linewidth}

\renewcommand{\epigraphflush}{center}
\vspace{-1mm}
\epigraph{Trust, but verify.}{\textit{Russian Proverb}}
\vspace{-2mm}

\section{Introduction}


Recent developments have enabled large language models (LLMs) to generate fluent text and follow instructions~\citep{wei2022finetuned, wang-etal-2023-self-instruct, NEURIPS2022_b1efde53, openai2023gpt4}. However, LLMs are known to produce seemingly plausible but erroneous outputs, often referred to as hallucinations~\citep[\textit{i.a.}]{ji2022survey, Zhang2023SirensSI}. This poses significant risks to downstream users due to the difficulty of fact-checking seemingly convincing generations from LLMs~\citep{Yue2023AutomaticEO, min2023factscore, Asai2024ReliableAA}.
One of the important desiderata for LLMs is thus \textit{verifiability}, the ability to ground responses to supporting evidence and render the produced claims easy to verify for humans. Verifiability allows users to uncover the competency of LLMs and \textit{calibrate} user trust, a crucial aspect of building trustworthy human-machine relationships~\citep{MUIR1987527}. 



Recent work increases verifiability through external artifacts such as producing citations~\citep{menick2022teaching, gao-etal-2023-enabling}, retrieving documents~\citep{NEURIPS2020_6b493230},
or post-hoc attribution methods~\citep{han2022orca}. 
Although helpful, these approaches provide no guarantee of relevance or usefulness. Models generations can be unfaithful to the retrieved documents in the context~\citep{shi2023trusting}, generative search engines often produce citations that are irrelevant or inaccurate~\citep{liu2023evaluating}, and explanations alone do not lead to verifiability~\citep{https://doi.org/10.1002/aaai.12182}.


We overcome the windingness of previous approaches through a \textit{verifiable-by-design} approach: generating \textbf{verbatim quotes} from high-quality sources such as Wikipedia. By determining verbatim quotes  from large-scale and high-quality corpora with efficient membership testing tools 
\citep{marone2023dataportraits}, those generations provide a natural method for attributing and verifying the correctness of generated claims.



\begin{figure*}[t]
    \centering
    \includegraphics[width=\textwidth]{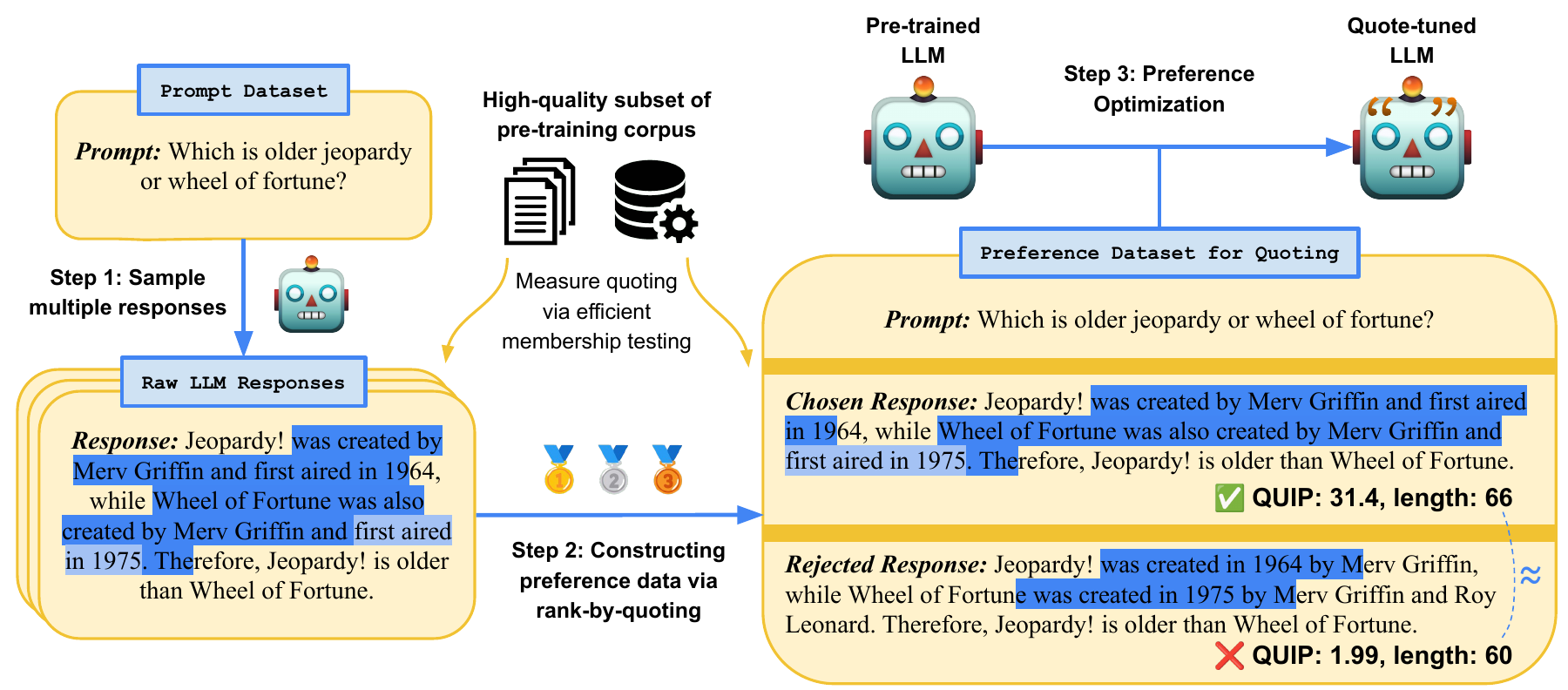}
    \caption{Pipeline of \method. The algorithm works by (1) sampling multiple responses from a pre-trained LLM, (2) constructing preference data via rank-by-quoting, and (3) preference optimization to quote.
    } 
    \label{fig:pipeline}
    \vspace{-3mm}
\end{figure*}

LLMs' potential to quote is driven by the observation that they are pre-trained on internet scale data---a subset of which contains high quality, reliable information---and that they have memorized a wide range of content from the pre-training stage~\citep{carlini21extracting, carlini2022quantifying, NEURIPS2023_59404fb8, Hartmann2023SoKMI}. 
Such analyses focus on \textit{covert} memorization and use adversarial prompts to extract the memorized contents~\citep{carlini21extracting, nasr2023scalable}. 
However, it remains an open question whether one can adapt LLMs to utilize their parametric knowledge to generate \textit{contextual} quotations 
across \emph{a wide range of} input prompts (not just  specialized or adversarial ones) on {realistic tasks} that require long-form generation.

We show this is indeed feasible with \method, our proposed method that aligns LLMs to quote through preference optimization and automatic feedback, without the need for human annotation. 
\method\ first generates responses from a pre-trained LLM, and then synthesizes a preference dataset for quoting by ranking responses by how much they quote from a desired high-quality corpus, e.g., Wikipedia.
Finally, \method\ aligns the model to quote from trusted sources by applying preference optimization algorithms (e.g., direct preference optimization~\citep{rafailov2023direct}) on the synthesized reference dataset. Figure \ref{fig:pipeline} illustrates the three-staged ``generate, synthesize, then tune" pipeline of \method.

Experiment results on long-form QA and open-ended text completion show that \method\ significantly increases quoting by up to 130\% relative to base models 
while maintaining or improving downstream performance (\Sref{sec:exp}). 
Moreover, \method{} generalizes to other domains and diverse model families, 
and enhances the truthfulness as measured by TruthfulQA~\citep{lin-etal-2022-truthfulqa} (\Sref{subsec:tqa}).

In summary, we present \method, a simple but effective technique for aligning LLMs to quote from their pre-training data. The quoted responses are \textit{verifiable-by-design} by inducing better verifiability without the need for human annotation and external knowledge bases (only leveraging parametric knowledge). \method\ sheds light on the feasibility of directly aligning language models to quote for trustworthiness, complementary to relying on non-parametric knowledge bases.

\section{Preliminaries}


\paravs
\paragraph{Quantifying Quoting}
We define a text string $x$ as \textit{quoted} from a corpus $C$ if a verbatim copy of $x$ is contained in $C$.
This design allows us to use \dataportrait~\citep{marone2023dataportraits}, a membership testing tool based on Bloom Filters~\citep{bloom1970spacetime}, to efficiently check whether text n-grams have appeared in the corpus. 
Specifically, we use Quoted Information Precision Score (QUIP-Score) metric proposed by \citet{weller2024accordingto}: 
\begin{equation*}
    \text{QUIP}_C(x) = \frac{\sum_{\text{gram}_n\in x}\mathbbm{1}_C(\text{gram}_n)}{|\text{gram}_n\in x|},
\end{equation*}
where $x$ is a text string, and $C$ is a trusted corpus,  
$\text{gram}_n\in x$ indicates all n-grams in $x$, and $\mathbbm{1}_C(\cdot)$ is an indicator function implemented by \textsc{Data Portraits} that return 1 if $\text{gram}_n\in C$ else 0. Intuitively, $\text{QUIP}_C(x)$ measures the percentage of n-grams in $x$ that appeared in $C$.\footnote{We follow the original implementation and use character 25-gram unless otherwise specified.}

\begin{algorithm*}[ht]
  \caption{\method 
    \label{alg:method}}
  \begin{algorithmic}[1]
    \Require{LLM policy $\pipt$, prompt dataset $\mathcal{D}_\text{prompt}=\{x^{(i)}\}_{i=1}^N$,  QUIP on corpus $C$, $\text{QUIP}_C(\cdot)$, QUIP hyperparameter $\delta_\text{quip}$, tokenized len2gth $len(\cdot)$, length hyperparameter $\delta_\text{length}$}
    \Ensure{Quoting-aligned LLM policy $\pi_\theta$}
    \Statex
    \State \hl{\texttt{//Sample Responses + Synthesizing Data}}
    \Let{$\mathcal{D}$}{$\emptyset$}
    \For{$i = 1,\dots, N$}
        \State $y_1,\dots,y_T\sim \pipt(\cdot | x^{(i)})$ \Comment{Sample responses from LLM policy}
        \Let{$\tilde{y}_1,\dots,\tilde{y}_T$}{\texttt{sort}$(y_1,\dots,y_T; \lambda y.-\text{QUIP}_C(y))$ \label{line:sort}} \Comment{Sort by decreasing QUIP order}
        \For{$w\in\{1,\dots,T-1\}$, $l\in\{w+1,\dots,T\}$} 
            \If{{$\text{QUIP}_C(\tilde{y}_w) - \text{QUIP}_C(\tilde{y}_l) > \delta_\text{quip}$ \textbf{and} $\frac{|len(\tilde{y}_w)-len(\tilde{y}_l)|}{\min\{len(\tilde{y}_w),~len(\tilde{y}_l)\}}<\delta_\text{length}$}}
                \Let{$\mathcal{D}$}{$\mathcal{D}\cup\{(x^{(i)},\tilde{y_w},\tilde{y}_l)\}$} 
                \State \textbf{break} 
            \EndIf
        \EndFor
    \EndFor
    \State \hl{\texttt{//Preference Optimization}}
    \State Initialize $\pi_\theta=\pipt$, and fine-tune $\pi_\theta$ on $\mathcal{D}$ using $\mathcal{L}_\text{DPO}$. 
    \State \textbf{return} $\pi_\theta$
  \end{algorithmic}
\end{algorithm*}

\paravs
\paragraph{Preference Optimization}
We review direct preference optimization \citep[DPO;][]{rafailov2023direct}, an algorithm for optimizing human preferences without reinforcement learning. Given a pre-trained LLM policy $\pipt$ and prompt $x$, a pair of responses $(y_1,y_2)\sim\pi_\text{ef}(\cdot | x)$ is sampled from the pre-trained model. 
The response pair is subsequently evaluated for preference (human annotators or automated metrics), with the more favored response labeled as \( y_w \) and the less preferred one as \( y_l \).
DPO assumes a static pairwise preference dataset $\mathcal{D}=\{x^{(i)}, y_w^{(i)}, y_l^{(i)}\}_{i=1}^N$. The loss function for optimizing the parameterized LLM policy $\pi_\theta$ is the following likelihood objective:
\begin{flalign*}
    &\mathcal{L}_\text{DPO}(\pi_\theta;\pipt) = 
    -\mathbb{E}_{(x,y_w,y_l)\sim\mathcal{D}}
    \\
    &\Big[
        \log\sigma\Big(
            \beta\log\frac{\pi_\theta(y_w|x)}{\pipt(y_w|x)}
            -\beta\log\frac{\pi_\theta(y_l|x)}{\pipt(y_l|x)}
        \Big)
    \Big],
\end{flalign*}
where $\pi_\theta$ is initialized as $\pipt$, $\sigma$ is the sigmoid function, and $\beta$ is a hyperparameter.

\section{Aligning LLMs to Quote with \method}
\label{sec:method}


The design of \method\ is inspired by the observation that preference datasets can be used to elicit desired behaviors in LMs using preference alignment frameworks~\citep{NIPS2017_d5e2c0ad, ziegler2019fine, NEURIPS2022_b1efde53}. 
The prior work, for example, has used this approrach to address  
factuality~\citep{tian2024finetuning}, honesty~\citep{Yang2023AlignmentFH}, harmlessness~\citep{bai2022constitutional, languagebarrier2024shen}, and relevance~\citep{wu2023finegrained}. We investigate whether automatic measures of quoting can be used to align LLMs to \textit{quote} from their pre-training data. 
We introduce our methodology here and empirically show its feasibility in \Sref{sec:exp}.


Illustrated in Alg.~\ref{alg:method}, \method\ works by sampling multiple responses from the to-be-tuned model, synthesizing preference pairs for quoting, and preference optimization. We now detail each step.
First, given a pre-trained LLM policy $\pipt$, for each prompt $x^{(i)}$ in a prompt dataset $\mathcal{D}_\text{prompt}$, we sample $T$ responses $y_1,\dots,y_T\sim \pipt(\cdot | x^{(i)})$ from the policy. Next, we construct pairwise preference data $(x^{(i)},y_w,y_l)$ by selecting a pair of response $(y_w,y_l)$ (where $y_w$ is more preferred) from $y_1,\dots,y_T$ that satisfies two constraints:

\paravs
\paragraph{Constraint 1: quoting.} $\text{QUIP}_C(y_w) - \text{QUIP}_C(y_l) > \delta_\text{quip}$, where $\delta_\text{quip}>0$ is a hyperparameter. Core to \method, this constraint ensures that the preferred response is more quoted than the dispreferred one.

\paravs
\paragraph{Constraint 2: length.} $\frac{|len(y_w)-len(y_l)|}{\min\{len(y_w),~len(y_l)\}}<\delta_\text{length}$, where $\delta_\text{length}\in(0,1)$ is a hyperparameter. Motivated by recent findings that RLHF and direct preference optimization approaches lead to increased response length~\citep{Singhal2023ALW, dubois2023alpacafarm}, we regularize the preferred and dispreferred responses to have similar tokenized length with each other. We provide an ablation of the length constraint in \Sref{appsec:length_constraint}.

\begin{table*}[ht]
    \centering
    \small
    \begin{tabular}{p{408pt}c}
        \toprule
        \textbf{\textit{Question:}} Who won the most MVP awards in the NBA? & \textbf{QUIP}\\

        \midrule

        \textbf{Reference}: 
        \quoteda{E}\quoteda{v}\quoteda{e}\quoteda{r}\quoteda{y}\quoteda{ }\quoteda{p}\quoteda{l}\quoteda{a}\quoteda{y}\quoteda{e}\quoteda{r}\quoteda{ }\quoteda{w}\quoteda{h}\quoteda{o}\quoteda{ }\quoteda{h}\quoteda{a}\quoteda{s}\quoteda{ }\quoteda{w}\quoteda{o}\quoteda{n}\quoteda{ }\quoteda{t}\quoteda{h}\quoteda{i}\quoteda{s}\quoteda{ }\quoteda{a}\quoteda{w}\quoteda{a}\quoteda{r}\quoteda{d}\quoteda{ }\quoteda{a}\quoteda{n}\quoteda{d}\quoteda{ }\quoteda{h}\quoteda{a}\quoteda{s}\quoteda{ }\quoteda{b}\quoteda{e}\quoteda{e}\quoteda{n}\quoteda{ }\quoteda{e}\quoteda{l}\quoteda{i}\quoteda{g}\quoteda{i}\quoteda{b}\quoteda{l}\quoteda{e}\quoteda{ }\quoteda{f}\quoteda{o}\quoteda{r}\quoteda{ }\quoteda{t}\quoteda{h}\quoteda{e}\quoteda{ }\quoteda{N}\quoteda{a}\quoteda{i}\quoteda{s}\quoteda{m}\quoteda{i}\quoteda{t}\quoteda{h}\quoteda{ }\quoteda{M}\quoteda{e}\quoteda{m}\quoteda{o}\quoteda{r}\quoteda{i}\quoteda{a}\quoteda{l}\quoteda{ }\quoteda{B}\quoteda{a}\quoteda{s}\quoteda{k}\quoteda{e}\quoteda{t}\quoteda{b}\quoteda{a}\quoteda{l}\quoteda{l}\quoteda{ }\quoteda{H}\quoteda{a}\quoteda{l}\quoteda{l}\quoteda{ }\quoteda{o}\quoteda{f}\quoteda{ }\quoteda{F}\quoteda{a}\quoteda{m}\quoteda{e}\quoteda{ }\quoteda{h}\quoteda{a}\quoteda{s}\quoteda{ }\quoteda{b}\quoteda{e}\quoteda{e}\quoteda{n}\quoteda{ }\quoteda{i}\quoteda{n}\quoteda{d}\quoteda{u}\quoteda{c}\quoteda{t}\quoteda{e}\quoteda{d}\quoteda{.}\quoteda{ }\quoteda{K}\quoteda{a}\quoteda{r}\quoteda{e}\quoteda{e}\quoteda{m}\quoteda{ }\quoteda{A}\quoteda{b}\quoteda{d}\quoteda{u}\quoteda{l}\quoteda{-}\quoteda{J}\quoteda{a}\quoteda{b}\quoteda{b}\quoteda{a}\quoteda{r}\quoteda{ }\quoteda{w}\quoteda{o}\quoteda{n}\quoteda{ }\quoteda{t}\quoteda{h}\quoteda{e}\quoteda{ }\quoteda{a}\quoteda{w}\quoteda{a}\quoteda{r}\quoteda{d}\quoteda{ }\quoteda{a}\quoteda{ }\quoteda{r}\quoteda{e}\quoteda{c}\quoteda{o}\quoteda{r}\quoteda{d}\quoteda{ }\quoteda{s}\quoteda{i}\quoteda{x}\quoteda{ }\quoteda{t}\quoteda{i}\quoteda{m}\quoteda{e}\quoteda{s}\quotedb{.}\quotedb{ }\quoteda{B}\quoteda{o}\quoteda{t}\quoteda{h}\quoteda{ }\quoteda{B}\quoteda{i}\quoteda{l}\quoteda{l}\quoteda{ }\quoteda{R}\quoteda{u}\quoteda{s}\quoteda{s}\quoteda{e}\quoteda{l}\quoteda{l}\quoteda{ }\quoteda{a}\quoteda{n}\quoteda{d}\quoteda{ }\quoteda{M}\quoteda{i}\quoteda{c}\quoteda{h}\quoteda{a}\quoteda{e}\quoteda{l}\quoteda{ }\quoteda{J}\quoteda{o}\quoteda{r}\quoteda{d}\quoteda{a}\quoteda{n}\quoteda{ }\quoteda{w}\quoteda{o}\quoteda{n}\quoteda{ }\quoteda{t}\quoteda{h}\quoteda{e}\quoteda{ }\quoteda{a}\quoteda{w}\quoteda{a}\quoteda{r}\quoteda{d}\quoteda{ }\quoteda{f}\quoteda{i}\quoteda{v}\quoteda{e}\quoteda{ }\quoteda{t}\quoteda{i}\quoteda{m}\quoteda{e}\quoteda{s}\quoteda{,}\quoteda{ }\quoteda{w}\quoteda{h}\quoteda{i}\quoteda{l}\quoteda{e}\quoteda{ }\quoteda{W}\quoteda{i}\quoteda{l}\quoteda{t}\quoteda{ }\quoteda{C}\quoteda{h}\quoteda{a}\quoteda{m}\quoteda{b}\quoteda{e}\quoteda{r}\quoteda{l}\quoteda{a}\quoteda{i}\quoteda{n}\quoteda{ }\quoteda{a}\quoteda{n}\quoteda{d}\quoteda{ }\quoteda{L}\quoteda{e}\quoteda{B}\quoteda{r}\quoteda{o}\quoteda{n}\quoteda{ }\quoteda{J}\quoteda{a}\quoteda{m}\quoteda{e}\quoteda{s}\quoteda{ }\quoteda{w}\quoteda{o}\quoteda{n}\quoteda{ }\quoteda{t}\quoteda{h}\quoteda{e}\quoteda{ }\quoteda{a}\quoteda{w}\quoteda{a}\quoteda{r}\quoteda{d}\quoteda{ }\quoteda{f}\quoteda{o}\quoteda{u}\quoteda{r}\quoteda{ }\quoteda{t}\quoteda{i}\quoteda{m}\quoteda{e}\quoteda{s}\quoteda{.}\quoteda{ }\quoteda{R}\quoteda{u}\quoteda{s}\quoteda{s}\quoteda{e}\quoteda{l}\quoteda{l}\quoteda{ }\quoteda{a}\quoteda{n}\quoteda{d}\quoteda{ }\quoteda{J}\quoteda{a}\quoteda{m}\quoteda{e}\quoteda{s}\quoteda{ }\quoteda{a}\quoteda{r}\quoteda{e}\quoteda{ }\quoteda{t}\quoteda{h}\quoteda{e}\quoteda{ }\quoteda{o}\quoteda{n}\quoteda{l}\quoteda{y}\quoteda{ }\quoteda{p}\quoteda{l}\quoteda{a}\quoteda{y}\quoteda{e}\quoteda{r}\quoteda{s}\quoteda{ }\quoteda{t}\quoteda{o}\quoteda{ }\quoteda{h}\quoteda{a}\quoteda{v}\quoteda{e}\quoteda{ }\quoteda{w}\quoteda{o}\quoteda{n}\quoteda{ }\quoteda{t}\quoteda{h}\quoteda{e}\quoteda{ }\quoteda{a}\quoteda{w}\quoteda{a}\quoteda{r}\quoteda{d}\quoteda{ }\quoteda{f}\quoteda{o}\quoteda{u}\quoteda{r}\quoteda{ }\quoteda{t}\quoteda{i}\quoteda{m}\quoteda{e}\quoteda{s}\quoteda{ }\quoteda{i}\quoteda{n}\quoteda{ }\quoteda{f}\quoteda{i}\quoteda{v}\quoteda{e}\quoteda{ }\quoteda{s}\quoteda{e}\quoteda{a}\quoteda{s}\quoteda{o}\quoteda{n}\quoteda{s}\quoteda{.} ...
        & 94.1\\
        \cdashlinelr{1-2}
        \textsc{\textbf{Llama2-7B-Chat}}: 
        \quoteda{L}\quoteda{e}\quoteda{B}\quoteda{r}\quoteda{o}\quoteda{n}\quoteda{ }\quoteda{J}\quoteda{a}\quoteda{m}\quoteda{e}\quoteda{s}\quoteda{ }\quoteda{h}\quoteda{a}\quoteda{s}\quoteda{ }\quoteda{w}\quoteda{o}\quoteda{n}\quoteda{ }\quoteda{t}\quoteda{h}\quoteda{e}\quoteda{ }\quoteda{m}ost NBA MVP awards with 4\quoteda{ }\quoteda{c}\quoteda{h}\quoteda{a}\quoteda{m}\quoteda{p}\quoteda{i}\quoteda{o}\quoteda{n}\quoteda{s}\quoteda{h}\quoteda{i}\quoteda{p}\quoteda{s}\quoteda{,}\quoteda{ }\quoteda{f}\quoteda{o}\quoteda{l}\quoteda{l}\quoteda{o}\quoteda{w}\quoteda{e}\quoteda{d}\quoteda{ }\quoteda{c}losely by Bill Russell with 5 MVP awards and 11 championships. Michael\quoteda{ }\quoteda{J}\quoteda{o}\quoteda{r}\quoteda{d}\quoteda{a}\quoteda{n}\quoteda{ }\quoteda{i}\quoteda{s}\quoteda{ }\quoteda{a}\quoteda{l}\quoteda{s}\quoteda{o}\quoteda{ }\quoteda{a}\quoteda{m}\quoteda{o}\quoteda{n}\quoteda{g}\quoteda{ }\quoteda{t}\quoteda{h}\quoteda{e}\quoteda{ }\quoteda{t}\quoteda{o}\quoteda{p}\quoteda{ }MVP winners with 6 championships and 5 MVP awards.
        & 3.9\\
        
        \cdashlinelr{1-2}
        \textbf{\method}: 
        \quoteda{T}\quoteda{h}\quoteda{e}\quoteda{ }\quoteda{p}\quoteda{l}\quoteda{a}\quoteda{y}\quoteda{e}\quoteda{r}\quoteda{ }\quoteda{w}\quoteda{h}\quoteda{o}\quoteda{ }\quoteda{h}\quoteda{a}\quoteda{s}\quoteda{ }\quoteda{w}\quoteda{o}\quoteda{n}\quoteda{ }\quoteda{t}\quoteda{h}\quoteda{e}\quoteda{ }\quoteda{m}\quoteda{o}\quotedb{s}\quotedb{t}\quotedb{ }\quoteda{M}\quoteda{o}\quoteda{s}\quoteda{t}\quoteda{ }\quoteda{V}\quoteda{a}\quoteda{l}\quoteda{u}\quoteda{a}\quoteda{b}\quoteda{l}\quoteda{e}\quoteda{ }\quoteda{P}\quoteda{l}\quoteda{a}\quoteda{y}\quoteda{e}\quoteda{r}\quoteda{ }\quoteda{(}\quoteda{M}\quoteda{V}\quoteda{P}\quoteda{)}\quotedb{ }\quotedb{a}\quotedb{w}\quotedc{a}\quotedc{r}\quotedc{d}\quotedc{s}\quotedc{ }\quotedb{i}\quotedb{n}\quotedb{ }\quotedb{t}\quotedb{h}\quotedb{e}\quotedb{ }\quotedb{N}\quotedb{a}\quotedb{t}\quotedb{i}\quotedb{o}\quotedb{n}\quotedb{a}\quotedb{l}\quotedb{ }\quotedb{B}\quotedb{a}\quoteda{s}\quoteda{k}\quoteda{e}\quoteda{t}\quoteda{b}\quoteda{a}\quoteda{l}\quoteda{l}\quoteda{ }\quoteda{A}\quoteda{s}\quoteda{s}\quoteda{o}\quoteda{c}\quoteda{i}\quoteda{a}\quoteda{t}\quoteda{i}\quoteda{o}\quoteda{n}\quoteda{ }\quoteda{(}\quoteda{N}\quoteda{B}\quoteda{A}\quoteda{)}\quotedb{ }\quotedb{i}\quotedb{s}\quotedb{ }\quoteda{K}\quoteda{a}\quoteda{r}\quoteda{e}\quoteda{e}\quoteda{m}\quoteda{ }\quoteda{A}\quoteda{b}\quoteda{d}\quoteda{u}\quoteda{l}\quoteda{-}\quoteda{J}\quoteda{a}\quoteda{b}\quoteda{b}\quoteda{a}\quotedb{r}\quotedb{,}\quotedb{ }\quotedb{w}\quotedb{h}\quotedb{o}\quotedb{ }\quotedb{h}\quotedb{a}\quotedb{s}\quotedb{ }\quoteda{w}\quoteda{o}\quoteda{n}\quoteda{ }\quoteda{t}\quoteda{h}\quoteda{e}\quoteda{ }\quoteda{a}\quoteda{w}\quoteda{a}\quoteda{r}\quoteda{d}\quotedb{ }\quotedb{a}\quotedb{ }\quotedb{r}\quotedb{e}\quotedb{c}\quotedb{o}\quotedb{r}\quotedb{d}\quotedb{ }\quotedb{s}\quotedb{i}\quotedb{x}\quotedb{ }\quotedb{t}\quotedb{i}\quotedb{m}\quotedb{e}\quotedb{s}\quotedb{ }\quoteda{d}\quoteda{u}\quoteda{r}\quoteda{i}\quoteda{n}\quoteda{g}\quoteda{ }\quoteda{h}\quoteda{i}\quoteda{s}\quoteda{ }\quoteda{c}\quoteda{a}\quoteda{r}\quoteda{e}\quoteda{e}\quoteda{r}\quoteda{ }\quoteda{w}\quoteda{i}\quoteda{t}\quoteda{h}\quoteda{ }\quoteda{t}\quoteda{h}\quoteda{e}\quoteda{ }\quoteda{M}\quoteda{i}\quoteda{l}\quoteda{w}\quoteda{a}\quoteda{u}\quoteda{k}\quoteda{e}\quoteda{e}\quoteda{ }\quoteda{B}\quoteda{u}\quoteda{c}\quoteda{k}\quoteda{s}\quoteda{ }\quoteda{a}\quoteda{n}\quoteda{d}\quoteda{ }\quoteda{t}\quoteda{h}\quoteda{e}\quoteda{ }\quoteda{L}\quoteda{o}\quoteda{s}\quoteda{ }\quoteda{A}\quoteda{n}\quoteda{g}\quoteda{e}\quoteda{l}\quoteda{e}\quoteda{s}\quoteda{ }\quoteda{L}\quoteda{a}\quoteda{k}\quoteda{e}\quoteda{r}\quoteda{s}\quoteda{.}\quoteda{ }\quoteda{A}bdul-\quoteda{J}\quoteda{a}\quoteda{b}\quoteda{b}\quoteda{a}\quoteda{r}\quotedb{ }\quotedb{p}\quotedb{l}\quotedb{a}\quotedb{y}\quotedb{e}\quotedb{d}\quotedb{ }\quotedb{i}\quotedb{n}\quotedb{ }\quotedb{t}\quotedb{h}\quotedb{e}\quotedb{ }\quotedb{N}\quotedb{B}\quotedb{A}\quotedb{ }\quotedb{f}\quoteda{r}\quoteda{o}\quoteda{m}\quoteda{ }\quoteda{1}\quoteda{9}\quoteda{6}9 to\quoteda{ }\quoteda{1}\quoteda{9}\quoteda{8}\quoteda{9}\quoteda{ }\quoteda{a}\quoteda{n}\quoteda{d}\quoteda{ }\quoteda{i}\quoteda{s}\quoteda{ }\quoteda{w}\quoteda{i}\quoteda{d}\quoteda{e}\quoteda{l}\quoteda{y}\quoteda{ }\quoteda{c}\quoteda{o}\quoteda{n}\quoteda{s}\quoteda{i}\quoteda{d}\quoteda{e}\quoteda{r}\quoteda{e}\quoteda{d}\quoteda{ }\quoteda{o}\quoteda{n}\quoteda{e}\quoteda{ }\quoteda{o}\quoteda{f}\quoteda{ }\quoteda{t}\quoteda{h}\quoteda{e}\quoteda{ }\quoteda{g}\quoteda{r}\quoteda{e}\quoteda{a}\quoteda{t}\quoteda{e}\quoteda{s}\quoteda{t}\quoteda{ }\quoteda{b}\quoteda{a}\quoteda{s}\quoteda{k}\quoteda{e}\quoteda{t}\quoteda{b}\quoteda{a}\quoteda{l}\quoteda{l}\quoteda{ }\quoteda{p}\quoteda{l}\quoteda{a}\quoteda{y}\quoteda{e}\quoteda{r}\quoteda{s}\quoteda{ }\quoteda{o}\quoteda{f}\quoteda{ }\quoteda{a}\quoteda{l}\quoteda{l}\quoteda{ }\quoteda{t}\quoteda{i}\quoteda{m}\quoteda{e}\quoteda{.}
        & 60.6\\
        \bottomrule
    \end{tabular}
    \caption{Example outputs generated by \textsc{Llama2-7B-Chat} before and after \method\ on NQ. \quoteda{Highlighted} segments are quoted from Wikipedia that appeared in the Pile~\citep{gao2020pile}. \quotedb{Lighter highlighting} and \quotedc{lightest highlighting} indicates two or three overlapped quoted segments, respectively. The minimum length to be considered quoted is a character-level 25-gram match. \textbf{\method\ significantly improves quoting from Wikipedia.}} 
    \label{tab:example}
\end{table*}
If multiple pairs of responses satisfy the constraints, a single pair $(y_w,y_l)$ with the highest average QUIP-Score among the two responses will be selected.\footnote{The design to select a maximum of one response pair per prompt is to preserve the distribution of prompts. Prior work also experimented with employing all possible preference pairs~\citep{ouyang2022training, tian2024finetuning}, which we leave to future work.} This ensures the dispreferred response still maintains relatively high quoting. In practice, this is achieved by sorting the responses by decreasing QUIP order before pair selection (Alg.~\ref{alg:method}, line \ref{line:sort}). If no response pair can be selected, the prompt $x^{(i)}$ is discarded. 

The reason why we employ model self-generation as the preferred response, instead of simply using verbatim quotes (e.g., spans of Wikipedia text that contains the gold answer) is twofold: (1) Using self-generated responses keeps the synthesized preference data on-policy, which is crucial for preference optimization as shown in recent work~\citep{tajwar2024preference}. (2) Because quotes are a subset of the model response, they are constrained to be highly contextually relevant to both the query and the surrounding response context. This relevance would be harder to achieve with standalone quotes. 

Finally, having obtained the synthetic preference dataset for quoting $\mathcal{D}$, we conduct DPO using $\mathcal{D}$ on the pre-trained LLM policy $\pipt$ to obtain the quoting-aligned policy $\pi_\theta$.

\begin{table*}[t]
\centering

\scalebox{0.88}
{
\begin{tabular}{@{}c|lccccc@{}}
\toprule
\multicolumn{1}{c}{}& & \textbf{\textit{Quoting}} & \multicolumn{2}{c}{\textbf{\textit{Adequacy}}} & \multicolumn{1}{c}{\textbf{\textit{Fluency}}} \\
\cmidrule(l){3-3}\cmidrule(l){4-5} \cmidrule(l){6-6}  
\multicolumn{1}{c}{\textit{\textbf{Setting}}} & \multicolumn{1}{c}{\textbf{\textit{Method}}} & {{QUIP}}$\uparrow$ & {Rouge-L}$\uparrow$ & {BARTSc}$\uparrow$ & {PPL}$\downarrow$ & {Length} \\ \midrule
\multirow{4}{*}{\begin{tabular}[c]{@{}c@{}}\textbf{\textit{In-Domain}}\\ NQ\end{tabular}} & \textsc{LLama2-7B-Chat} & 34.9 & 22.4 & -3.99 & 4.96 & 115.9 \\
 & +\textit{According-to} prompting & 36.2 & 22.9 & -3.95 & 4.55 & 129.6 \\
 & +Best-of-32 QUIP rerank & 50.4 & 23.3 & -3.98 & 4.40 & 110.2 \\
 & +\method & \textbf{54.5} & \textbf{24.2} & \textbf{-3.93} & \textbf{3.78} & 117.6 \\ \midrule
\multirow{4}{*}{\begin{tabular}[c]{@{}c@{}}\textit{\textbf{Out-of-Domain}}\\ NQ $\rightarrow$ ELI5\end{tabular}} & \textsc{LLama2-7B-Chat} & 26.8 & \textbf{18.8} & -4.78 & 3.93 & 179.8 \\
 & +\textit{According-to} prompting & 28.0 & 18.3 & \textbf{-4.75} & 3.56 & 225.7 \\
 & +Best-of-32 QUIP rerank & 37.6 & 18.7 & -4.78 & 3.72 & 173.8 \\
 & +\method\ on NQ & \textbf{41.4} & 18.3 & -4.84 & \textbf{3.55} & 179.6 \\ \bottomrule
\end{tabular}
}
\caption{Results on Long-Form QA datasets. QUIP and Rouge-L are in percentages. \method\ significantly improves QUIP-Score over baselines in both in- and out-of-domain QA tasks, while maintaining a similar quality of predicted answers as measured by Rouge-L, BARTScore, and Perplexity.
}
\label{tab:quoting}
\end{table*}

\paravs
\paragraph{Desirability of Quoting} 
We show an example of the model generation before and after \method\ in Table \ref{tab:example} and highlight segments that are quoted verbatim from the Pile~\citep{gao2020pile} subset of Wikipedia along with the corresponding QUIP-Score. The quoted segments are determined by conducting membership inference on character-level 25-gram substrings of generated text with \dataportrait{}~\citep{marone2023dataportraits}. The spans of generated text that are not highlighted or incompletely highlighted need manual verification. \textbf{More quoting encouraged by \method\ leads to fewer spans that need to be verified and, thus, better verifiability.} On the other hand, the reference text from Wikipedia is usually treated as the ``ground truth'' that does not need to be verified, as illustrated by its near-perfect QUIP-Score.\footnote{The minor mismatch is due to preprocessing and potential version differences.} We provide an extended discussion of verifiability in \Sref{sec:related}.

We emphasize that the quality of the quoting corpus $C$ is important: $C$ needs to be carefully selected such that it contains high-quality, low-risk text, such as our choice of Wikipedia. We show in \Sref{subsec:tqa} that quoting from truthful corpus increase model truthfulness, and discuss implications of the selection of $C$ on privacy and copyright in \Sref{sec:discuss}.

Aside from better verifiability, \citet{weller2024accordingto} demonstrates that more quoting, as measured by QUIP-Score, leads to fewer hallucinations in the generated text. Our analysis in \Sref{subsec:tqa} shows that encouraging quoting leads to more truthful models. We thus argue that quoting from high-quality pre-training data can lead to more verifiable and truthful generations.



\section{Experiments}
\label{sec:exp}


In this section, we provide empirical evidence on how \method\ can provide better verifiability to LLM-generated responses, while maintaining generation quality. We conduct \method\ on the long-form QA (\Sref{subsec:lfqa}) and open-ended text completion (\Sref{subsec:textcomp}) tasks. Additionally, we show that quoting-aligned models are more truthful than their vanilla counterparts (\Sref{subsec:tqa}), and maintain downstream performance (\Sref{sec:downstream}).

\begin{table*}[t]
\centering

\scalebox{0.88}{
\begin{tabular}{lccccc}
\toprule
 & \textbf{\textit{Quoting}} & \multicolumn{2}{c}{\textbf{\textit{Adequacy}}} & \multicolumn{1}{c}{\textbf{\textit{Fluency}}} \\
\cmidrule(l){2-2}\cmidrule(l){3-4} \cmidrule(l){5-5}  
\multicolumn{1}{c}{\textbf{\textit{Method}}} & {{QUIP}}$\uparrow$ & {Rouge-L}$\uparrow$ & {BARTSc}$\uparrow$ & {PPL}$\downarrow$ & {Length} \\ \midrule
\textsc{LLama3.1-8B-Inst} & 33.0 & 22.5 & -3.97 & 5.03 & 136.7 \\
 +\method & \textbf{43.0} & \textbf{25.7} & \textbf{-3.82} & \textbf{3.13} & 118.3 \\
\midrule
\textsc{Gemma2-9B-IT} & 30.0 & 21.4 & -4.03 & 7.60 & 60.1 \\
 +\method & \textbf{44.9} & \textbf{24.4} & \textbf{-3.97} & \textbf{5.76} & 57.4 \\
\midrule
\textsc{Starling-7B-Beta} & 33.8 & 22.7 & -3.83 & 2.81 & 156.1 \\
 +\method & \textbf{44.4} & \textbf{23.8} & \textbf{-3.82} & \textbf{2.70} & 150.8 \\
 \bottomrule
\end{tabular}
}
\caption{\method{} on diverse model family consistently improves quoting, adequacy, and fluency over their respective base models on NQ.
}
\label{tab:quote_models}
\vspace{-2mm}
\end{table*}
\subsection{Improving Quoting in Long-Form QA}
\label{subsec:lfqa}

\paragraph{Task Construction~~~} In the long-form QA (LFQA) setting, we study whether \method\ can effectively increase quoting in model-generated answers given questions as the prompt. To find settings relevant to \method, we select datasets that induce \textit{long-form response} to measure quoting and also allow us to \textit{verify the answers from trusted sources} such as Wikipedia. Accordingly, we experiment on two datasets, NaturalQuestions \citep[NQ;][]{kwiatkowski-etal-2019-natural} and ELI5~\citep{fan-etal-2019-eli5}. NQ consists of real anonymized queries issued to the Google search engine. Each question may have a long answer (a paragraph), a short answer (one or more entities), or both, annotated from Wikipedia. We employ the subset of NQ that has long answers: we sample 20K training set questions to be used as the prompt dataset $\mathcal{D}_\text{prompt}$ for \method, and the full development set is used as the \textbf{in-domain} evaluation set. To evaluate whether quoting can be generalized to out-of-domain questions, we use the evaluation set of the ELI5 dataset, where questions are mined from the Reddit ``Explain Like I’m Five'' forum, as the \textbf{out-of-domain} evaluation set.
\paragraph{Baselines~~~} 
Aside from the pre-trained LLM policy $\pipt$, we consider the \textit{according-to} prompting method from \citet{weller2024accordingto}, which directs LLMs to ground responses against pre-training sources through prompting.\footnote{We use the best grounding prompt found in \citet{weller2024accordingto}, i.e., ``Respond to this question using only information that can be attributed to Wikipedia.''} Finally, we include a strong Best-of-N QUIP reranking baseline, where we sample 32 responses from the pre-trained model $\pipt$ and rerank the response by selecting the one with the highest QUIP-Score. Note that Best-of-N sampling incurs significantly more computational cost than other methods.\footnote{We also experimented with fine-tuning on NQ reference answers. However, we found this baseline ineffective and thus is omitted in the main results.}
\paragraph{Metrics~~~} 
To our main interest, we measure quoting with \textbf{QUIP-Score} using the Wikipedia subset of the Pile dataset~\citep{gao2020pile} as the grounding corpus $C$.\footnote{Although \textsc{Llama2} training data is not public, we believe Wikipedia is presumably observed in the pre-training corpus due to its widespread usage.} We report the \textbf{BARTScore}~\citep{yuan2021bartscore} and \textbf{Rouge-L}~\citep{lin2004rouge} between generated and reference answers as metrics for adequacy of generated answers. The perplexity (\textbf{PPL}) of generation text calculated by \textsc{Llama2-7B} is used as a measure for fluency. We also report average generation length as preference optimization could lead to length biases~\citep{Singhal2023ALW}.

\paragraph{Results}
\label{exp:id}
We run experiments with hyperparameters detailed in \Sref{appsec:training_details}. We first employ \textsc{Llama2-7B-Chat}~\citep{touvron2023llama2} as the pre-trained model $\pipt$. 
After DPO, the reward accuracy on a held-out evaluation set is 86.3\%, indicating that the model learns quoting preference reasonably well. For in-domain evaluation, we test \method\ against baselines on the evaluation set of NQ. 
Shown in Table \ref{tab:quoting} (upper), \textbf{\method\ significantly improves upon all baselines in quoting, even outperforming the strong Best-of-32 QUIP rerank baseline that is more computationally costly. }
In particular, \method\ enables a significant 56.2\% ($34.9\rightarrow54.5$) quoting improvement relative to the un-tuned \textsc{Llama2-7B-Chat} model.
\method\ also slightly improves answer adequacy and fluency. Because \method{} significantly increases quoting from Wikipedia, which contains high-quality text thoroughly curated by human editors, the responses from quote-tuned models benefit from the high-quality nature on Wikipedia.\footnote{Similar findings are also found in \citet{li2024nearestneighborspeculativedecoding}, which report improved perplexity by conducting speculative decoding to copy spans from a datastore.} While \textit{according-to} prompting slightly increases quoting at the expense of notably longer generation length, \method\ maintains similar answer length compared to \textsc{Llama2-7B-Chat} generations. An example output is available in Table \ref{tab:example}.

To test the out-of-domain generalization ability of \method, we use \textsc{Llama2-7B-Chat} quote-tuned on NQ for evaluation on ELI5. \method\ still outperforms all baselines in quoting, while maintaining adequacy and improving fluency compared to the original model. Table \ref{tab:quoting} (lower) shows that \textbf{\method\ generalizes quoting to out-of-domain prompts.} 

Finally, we apply \method{} on \textsc{Llama3.1}~\citep{dubey2024llama}, \textsc{Gemma2}~\citep{gemmateam2024gemma2improvingopen}, and \textsc{Starling}~\citep{zhu2024starlingb} models, and find \textbf{\method{} consistently improve quoting, adequacy, and fluency across diverse model families} (Table~\ref{tab:quote_models}).



\subsection{Improving Quoting in Open-Ended Text Completion}
\label{subsec:textcomp}



\paravs
\paragraph{Task Construction~~~} 
We now study whether \method\ can be applied to open-ended text completion, where we measure quoting on the candidate LLM's open-ended continuation of test prompts. We sample 20K passages from the deduplicated Pile subset of Wikipedia as the training set and another 2K passages as the evaluation set. For each passage, we use the first 32 tokens as the prompt and the remainder as the reference continuation, truncated to a maximum of 128 tokens. 

\paragraph{Baselines and Metrics~~~} We employ the pre-trained LLM policy $\pipt$ and Best-of-N QUIP reranking baselines following the LFQA setting (\Sref{subsec:lfqa}). Instead of according-to prompting, we fine-tune $\pipt$ on reference continuations of the train set as another baseline since $\pipt$ in this setting is not instruction-tuned.
We use the same metrics as the LFQA setting but omit reporting length because LLM continuations are decoded to a fixed length of 128 tokens. We use \textsc{Llama2-7B} as the pre-trained model $\pipt$, and measure perplexity with the \textsc{Mistral-7B} model instead to prevent self-evaluation bias~\citep{he-etal-2023-blind}.\footnote{Moreover, \textsc{Mistral-7B} is shown to be a stronger model~\citep{Jiang2023Mistral7} compared to \textsc{Llama2-7B}.}

{
\setlength{\tabcolsep}{3pt}
\begin{table}[ht]
\centering
\small

\resizebox{1\linewidth}{!}{
\begin{tabular}{lcccc}
\toprule

\multicolumn{1}{c}{} & \textbf{\textit{Quoting}} & \multicolumn{2}{c}{\textbf{\textit{Adequacy}}} & \multicolumn{1}{c}{\textbf{\textit{Fluency}}} \\
\cmidrule(l){2-2}\cmidrule(l){3-4} \cmidrule(l){5-5}  
\multicolumn{1}{c}{\textit{\textbf{Method}}} & {{QUIP}}$\uparrow$ & {R-L}$\uparrow$ & {BSc}$\uparrow$ & {PPL}$\downarrow$ \\ \midrule
\textsc{Llama2-7B} & 25.7 & 21.8 & -4.95 & 9.03 \\
+Fine-tuning & 29.1 & 21.9 & \textbf{-4.90} & 9.58 \\
+Bo32 QUIP rerank & 47.9 & \textbf{23.8} & -4.95 & 6.63 \\
+\method & \textbf{59.2} & 23.1 & -5.02 & \textbf{5.39} \\ \bottomrule
\end{tabular}
}
\caption{On the open-ended text completion setting, \method\ significantly improves quoting and fluency while maintaining adequacy.}
\label{tab:textcomp}
\end{table}
}

\begin{table*}[ht]
\setlength{\tabcolsep}{3.5pt} 
\centering
\small
\begin{tabular}{@{}llll|ll@{}}
\toprule
 & \multicolumn{3}{c|}{\textbf{Generation}} & \multicolumn{2}{c}{\textbf{Multiple Choice}} \\ \cmidrule(l){2-6} 
\multicolumn{1}{c}{\textbf{\textit{Method}}} & \multicolumn{1}{c}{{Truthful}} & \multicolumn{1}{c}{{Informative}} & \multicolumn{1}{c|}{Truthful$\times$Informative} & \multicolumn{1}{c}{MC1} & \multicolumn{1}{c}{MC2} \\ \midrule
\textsc{Llama2-7B-Chat} & 54.2 & \textbf{92.0} & 46.6 & 30.2 & 45.3 \\
+\method & \textbf{61.8 (+14.0\%)} & 89.5 (-2.7\%) & \textbf{51.5 (+10.5\%)} & \textbf{32.8 (+8.5\%)} & \textbf{47.9 (+5.6\%)} \\ \bottomrule
\end{tabular}
\caption{Results on TruthfulQA. \method\ improve model truthfulness even though not explicitly tuned for truthfulness, suggesting that quoting from pre-train data indirectly improves the truthfulness of generations.
}
\label{tab:tqa}
\end{table*}
\paravs
\paragraph{Results}
After conducting \method{} with hypermaraters detailed in \Sref{appsec:training_details}, the reward accuracy on a held-out evaluation set is $84.0\%$. As shown in Table \ref{tab:textcomp}, \textbf{\method\ significantly improves both quoting and fluency over all baselines}. Notably, \method\ more than doubles the QUIP-Score compared to the pre-trained \textsc{Llama2-7B} baseline ($25.7\rightarrow 59.2$, a 130.4\% relative increase), and outperforms the strong QUIP reranking baseline. On the other hand, \method\ maintains a similar adequacy of generated answers compared to \textsc{Llama2-7B}.

Interestingly, Table \ref{tab:textcomp} shows that simply reranking \textsc{Llama2-7B} generation by QUIP can lead to a better perplexity as measured by \textsc{Mistral-7B}.
We hypothesize that because Wikipedia is an encyclopedia that has been revised multiple times and contains mostly high-quality text, quoting from this canonical corpus also has benefits of fluency aside from better verifiability.




\section{Analysis}
\subsection{Effect of Quoting on Truthfulness}
\label{subsec:tqa}

We hypothesize that besides increasing verifiability, quoting from high-quality corpora such as Wikipedia might also increase truthfulness because LLMs are aligned to rely on trustworthy sources. To verify this hypothesis, we take the quote-tuned model from the LFQA setting (\Sref{subsec:lfqa}) and evaluate its performance on the TruthfulQA dataset~\citep{lin-etal-2022-truthfulqa}. We follow the standard evaluation procedure on TruthfulQA, which fine-tunes GPT-3~\citep{brown2020language} on human annotations as truthfulness and informativeness judges. We defer further details to Appendix \ref{appsec:tqa_eval}.

As shown in Table \ref{tab:tqa}, \method\ increases model truthfulness, as well as answers that are both truthful and informative, over the un-tuned \textsc{Llama2-7B-Chat} model by a notable margin. On the other hand, informativeness slightly dropped, suggesting that the quote-tuned model is more conservative and has an increased tendency to decline to answer. We provide example outputs in Table \ref{tab:tqa_example}. Interestingly, \textbf{\method\ can improve model truthfulness even though not explicitly tuned to do so}: because the preference optimized in \method\ is only quoting as measured by QUIP-Score, the model is not directly optimized to be factual, in contrast to works that directly aims at truthfulness or factuality~\citep{tian2024finetuning,li2023inferencetime}. The increase of truthfulness likely attributes to the fact that Wikipedia is a relatively high-quality and reliable source, and quoting more from such reliable sources leads to more truthful responses. We thus posit aligning LLMs to reliable sources is a promising approach to increase their truthfulness.



{
\subsection{Evaluation of Downstream Performance}
\label{sec:downstream}

Because \method\ trains model on specialized long-form generation tasks, it is an open question whether the significant increase of quoting would lead to degradation of general capabilities. Thus, we now test the model before and after quote-tuning on general capability benchmarks MMLU~\citep{hendrycks2020measuring}, GSM8K~\citep{cobbe2021gsm8k}, BIG-Bench Hard~\citep[BBH;][]{suzgun-etal-2023-challenging}, and Hellaswag~\citep[HS;][]{zellers2019hellaswag}.\footnote{We conduct evaluation using the \href{https://github.com/EleutherAI/lm-evaluation-harness}{lm-evaluation-harness} framework under default settings.}

\setlength{\tabcolsep}{2.4pt}
\begin{table}[ht]
\centering
\small
\resizebox{1\linewidth}{!}{
\begin{tabular}{lcccc}
\toprule
               & \textbf{MMLU}  & \textbf{GSM8K} & \textbf{BBH}  & \textbf{HS} \\ \midrule
\textsc{LLama2-7B-Chat} & 46.38 & 20.92  & 40.21 & 75.51     \\ 
+\method  & 45.65 & 19.79  & 39.47 & 73.96     \\ \midrule
$\Delta$         & -0.73 & -1.13  & -0.74 & -1.55     \\ \bottomrule
\end{tabular}
}
\caption{Evaluation on general capability benchmarks. \method\ only post minor degradation while significantly improve quoting.}
\label{tab:benchmark}
\end{table}
}

As shown in Table~\ref{tab:benchmark}, \method\ only leads to very small degradations (less than two points for all tested benchmarks), while significantly improving quoting. We therefore find \method{} a significantly better trade-off for verifiability with a small cost of general capability.

\begin{figure*}%
    \centering
    \subfloat{{\includegraphics[trim=0mm 0mm 0mm 0mm, clip, height=0.30\textwidth]{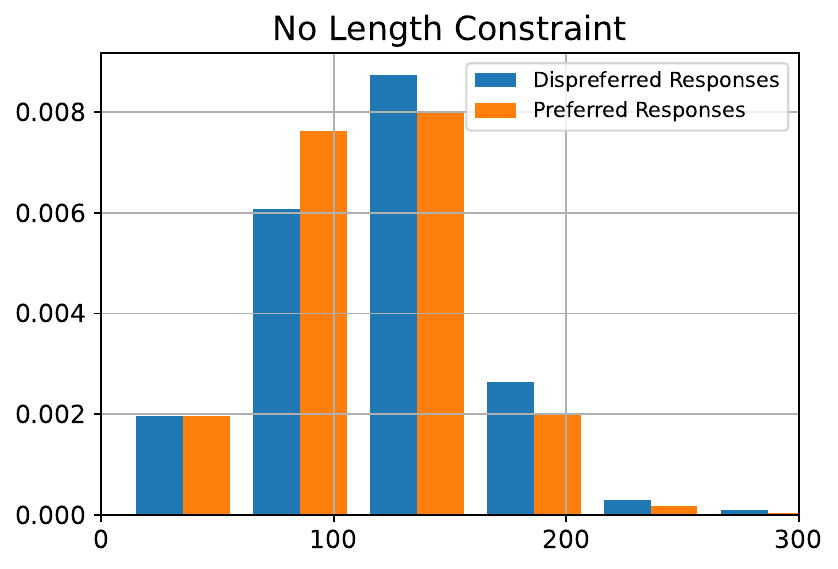} }}%
    \qquad
    \subfloat{{\includegraphics[trim=0mm 0mm 0mm 0mm, clip, height=0.30\textwidth]{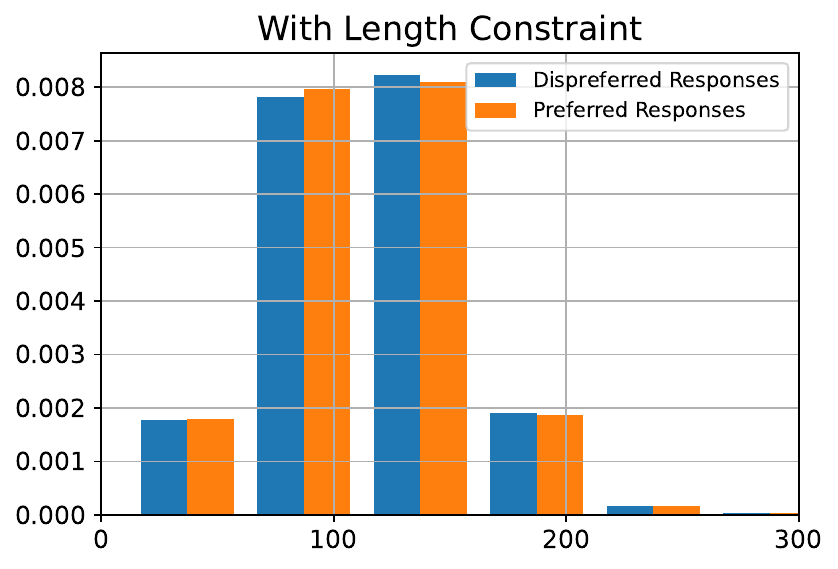} }}%
    \caption{Length distribution of the dispreferred and preferred responses with or without the length constraint on NQ. \textbf{Left}: No length constraint. \textbf{Right}: added length constraint with $\delta_\text{length}=0.1$. Adding length constraints properly regulates length distribution of responses.}%
    \label{fig:length_distribution}%
\end{figure*}

\begin{table*}[t]
\small
\centering
\scalebox{1}{
\begin{tabular}{@{}c|lccccc@{}}
\toprule
\multicolumn{1}{c}{}& & \textbf{\textit{Quoting}} & \multicolumn{2}{c}{\textbf{\textit{Adequacy}}} & \multicolumn{1}{c}{\textbf{\textit{Fluency}}} \\
\cmidrule(l){3-3}\cmidrule(l){4-5} \cmidrule(l){6-6}  
\multicolumn{1}{c}{\textit{\textbf{Setting}}} & \multicolumn{1}{c}{\textbf{\textit{Method}}} & {{QUIP}}$\uparrow$ & {Rouge-L}$\uparrow$ & {BARTSc}$\uparrow$ & {PPL}$\downarrow$ & {Length} \\ \midrule
\multirow{3}{*}{\begin{tabular}[c]{@{}c@{}}\textbf{\textit{In-Domain}}\\ NQ\end{tabular}} & \textsc{LLama2-7B-Chat} & 34.9 & 22.4 & -3.99 & 4.96 & 115.9 \\
 & +\method & \textbf{54.5} & {24.2} & \textbf{-3.93} & \textbf{3.78} & 117.6 \\ 
 & +\textsc{QT} w/o len. constraint & 53.6 & \textbf{24.4} & {-3.95} & {3.88} & 105.9 \\ 
 \midrule
\multirow{3}{*}{\begin{tabular}[c]{@{}c@{}}\textit{\textbf{Out-of-Domain}}\\ NQ $\rightarrow$ ELI5\end{tabular}} & \textsc{LLama2-7B-Chat} & 26.8 & \textbf{18.8} & \textbf{-4.78} & 3.93 & 179.8 \\
 & +\textsc{QT} on NQ & \textbf{41.4} & 18.3 & -4.84 & \textbf{3.55} & 179.6 \\
 & +\textsc{QT} on NQ w/o len. constraint & {40.5} & 18.6 & -4.85 & {3.84} & 154.3 \\ \bottomrule
\end{tabular}
}

\caption{Results on the ablation of the length constraint. \textsc{QT} is short for \method. Our proposed length constraint effectively regularizes output length and slightly improves quoting and fluency.}
\label{tab:length_ablation}
\vspace{1mm}
\end{table*}

\subsection{Ablation of the Length Constraint}
\label{appsec:length_constraint}
We conduct an ablation on the length constraint of the \method\ algorithm on the LFQA setting, relaxing the constraint that the preferred and dispreferred responses need to have similar lengths to each other. Experimental results are shown in Table~\ref{tab:length_ablation}. While \method\ leads to responses that have very similar lengths with the un-tuned model (117.6 vs 115.9 on NQ, 179.6 vs 179.8 on ELI5), \method\ without the length constraint leads to notably \textit{shorter} response (105.9 on NQ, 154.3 on ELI5). 

We hypothesize this phenomenon is due to the bias within synthetic preference data where length is not regularized: as shown in Figure~\ref{fig:length_distribution}, the density of preferred response is notably higher than dispreferred ones around length 100. We speculate that this is caused by the sampled responses having a non-uniform distribution of QUIP-Score over different length ranges, which we provide empirical evidence in Figure \ref{fig:quip_gain}.

On the other hand, ablating the length constraint leads to slightly lower quoting, relatively similar adequacy, and notably worse perplexity compared to the full \method\ algorithm, depicting the effectiveness of the length constraint.




\section{Related Work} 
\label{sec:related}

\paragraph{Improving Verifiability}
Hallucination in LLMs~\citep{ji2022survey, Zhang2023SirensSI, Mishra2024FinegrainedHD} has motivated approaches that improve the verifiability of LLM generations. 
Recent work on improving the verifiability of LLM generations relies on \textbf{external artifacts}.
One emerging trend is training LLMs to produce {citations} that support generated claims~\citep{menick2022teaching, gao-etal-2023-enabling, Huang2024TrainingLM}. While citations improve attribution, LLM can still hallucinate incorrect or irrelevant citations~\citep{liu2023evaluating}, which is non-trivial to verify. 
\citet{khalifa2024sourceaware} introduce intrinsic source citation to enable more faithful attribution to parametric knowledge, on a synthetic setting. \method{} is an approach complementary to citations that enhances verifiability. \citet{hennigen2024towards} finds that explicit symbolic references to structured conditioning data, such as JSON tables, lead to faster human verification, further motivating our approach of increasing verifiability through verbatim quotes. 

Retrieval-augmented generation~\citep[\textit{i.a.}]{guu2020retrieval, lewis2020retrieval, borgeaud2022improving, JMLR:v24:23-0037} allows fact-checking generation with the retrieved documents as supporting evidence. \citet{min-etal-2023-nonparametric} used retrieved tokens directly as generation, but is limited to the masked-filling setting with short spans of text. However, checking against retrieved documents is still non-trivial and there is no guarantee that generated text is completely faithful to these documents. On the other hand, our framework for quoting, based on \citet{marone2023dataportraits, weller2024accordingto}, makes the verification of quoted segments from fact bases trivial, given that the target model is capable of producing rich quotations after \method. Our work, which focuses on parametric knowledge, is also complementary to methods that rely on non-parametric knowledge bases. 






\paragraph{Impact of Preference Data}
The construction of pairwise preference data significantly impacts model behavior. \citet{tian2024finetuning} fine-tunes LLMs to be more factual by constructing preference data with automatic measures of factuality \citep{min2023factscore} and model confidence scores. \citet{Yang2023AlignmentFH} formalizes aligns LLMs with being honest by constructing pairwise data that prefers answers only when the model possesses relevant knowledge and abstains from answering otherwise. 
\citet{Yuan2024SelfRewardingLM} iteratively constructs preference data by prompting LLMs themselves for quality measurements. \citet{shi2023saferinstruct} automates preference data generation with LMs, utilizing instruction tuning and expert LMs to synthesize high-quality preference data. Our work also synthesizes pairwise data that give preference to the one that quotes more from a given corpus. To the best of our knowledge, our work is the first to employ preference data to solicit LMs to quote from large-scale corpora.

\paragraph{Memorization}
Works have demonstrated that LLMs memorize a significant portion of their pre-training data \citep{carlini21extracting, carlini2022quantifying, hu2022membership, ippolito-etal-2023-preventing, NEURIPS2023_59404fb8, Hartmann2023SoKMI}, and we can extract them by adversarial prompting \citep{carlini21extracting, nasr2023scalable}. Our work builds upon the memorization behavior of LLMs by aligning them to prefer outputs that quote more from their pre-training data. Also related to our work, $k$NN-LMs \citep{khandelwal2019generalization} improve generalization by using nearest neighbor search to retrieve similar contexts from a datastore. 
We defer further related work to \Sref{appsec:related}.



\section{Discussions and Future Work}
\label{sec:discuss}






In \Sref{sec:exp}, we provide rich empirical evidence that \method{} can significantly promote parametric quoting across diverse tasks, domains, and model families. Since quoted texts are autoregressively generated as spans of the model responses, and our evaluation quantitatively shows that these responses are of high adequacy and fluency (Tables~\ref{tab:quoting},~\ref{tab:quote_models},~\ref{tab:textcomp}), generated quotes are therefore constrained to be highly adequate, fluent, and context-relevant text. 
These findings demonstrate that it is not only possible but easily feasible to leverage parametric knowledge to generate more verifiable outputs. We thus argue \textbf{current LLMs have abundant underutilized potential in improving their own verifiability}, and call for future work that develop more attributable, verifiable approaches through our proposed ``trivializing verification through quoting'' framework.

As an early exploration of the quoting framework, this work focuses on investigating the \textit{\textbf{feasibility}} of unlocking parametric quoting through fine-tuning, and the \textit{\textbf{generalizability}} of enhanced quoting across different domains, tasks, and model families. We have provided positive empirical evidence supporting both aspects. Therefore, our setup focuses on measuring and improving the overall rate of quoting, as evaluated by QUIP-Score, leaving room for future work to enhance quoting reward signals by considering other desiderata of quoted texts, such as quote completeness and usefulness under the current context. We provide an extended discussion of limitations in the subsequent section.

Quoting has implications on privacy, security, and copyright. We focus on enhancing quoting from Wikipedia, a high-quality and low-risk corpus, for verifiability. As pre-training data contains diverse mixtures of data with varying risk levels~\citep{elazar2024whatsbigdata, longpre-etal-2024-pretrainers}, we argue that the grounding corpus for \method{} must be carefully selected and limited to trusted, public sources such as Wikipedia to prevent privacy violations or copyright infringement.

In conclusion, our approach presents a promising direction for leveraging the parametric knowledge of LLMs to facilitate easier verification of model generation and improve the calibration of human-machine trust.

\section*{Limitations}
(i) Our work maximizes the amount of quoting measured by QUIP-Score~\citep{weller2024accordingto}, but does not distinguish between many short quotes v.s. a few long ones, where the latter is more preferable. 
Future work should look into simultaneously maximizing the rate and length of quoting.
(ii) Another future direction involves extending the experiments to other settings, such instruction tuning~\citep[\textit{i.a.}]{mishra2022cross,wei2021finetuned, wang-etal-2022-super, wang-etal-2023-self-instruct, Zhang2023InstructionTF}, where a diverse set of tasks are present. 
(iii) We explored quoting as an interface for parametric knowledge only. 
This leaves room for investigating the synergy between quote-tuned models and retrieval-augmented generation~\citep[\textit{i.a.}]{guu2020retrieval, lewis2020retrieval, borgeaud2022improving, JMLR:v24:23-0037} or other non-parametric techniques~\citep{min-etal-2023-nonparametric}. 
(iv) Finally, quoting provides a natural interface for attribution~\citep{Bohnet2022AttributedQA, muller-etal-2023-evaluating, Malaviya2023ExpertQAEQ, Slobodkin2024AttributeFT}. Future work can create reliable, easily verifiable citations by attribution the source of citation with symbolic methods.

\section*{Ethical Considerations}

We have shown that increasing verbatim quotes from pre-training data through \method{} is a promising approach for enhancing verifiability. However, the ability to quote from pre-training data have broad ethical implications. While enhanced quoting increase users' ability to attribute generation back to their original sources, this could be a \textit{double-edged sword} regarding privacy protection: adversarial users might utilize similar methods to extract sensitive information
contained in pre-training data. On the other hand, if this is feasible, it may create a path for auditing pre-training data.

Furthermore, the legal implications of quoting from pre-training data must be carefully managed. It is necessary to ensure that copyrighted material is handled appropriately, with proper attribution, and that the use of such data adheres to intellectual property laws. By addressing ethical and legal considerations, we envision \method{} as a responsible tools for enhancing verifiability and calibrating human-machine trust.



\section*{Acknowledgements}
This work is supported by ONR grant (N00014-24-1-2089) and an Amazon Faculty Research Award. 
We sincerely thank Yung-Sung Chuang, Andrew Wang, Jeffrey Cheng, Mahsa Yarmohammadi, Dongwei Jiang, and the broader JHU CLSP community for discussions and inspiration.





\bibliography{bib/anthology, bib/ref_feb5}

\appendix

\onecolumn

\begin{center}
{\Large \textbf{Supplemental Material}}
\end{center}

\section{Additional Related Work}
\label{appsec:related}
\paragraph{Reward Modeling and Preference Optimization} 
Works that align LMs to human preferences \citep{ziegler2019fine, NEURIPS2020_1f89885d, NEURIPS2022_b1efde53, bai2022training} train reward model on pairwise human preference data and use reinforcement learning algorithms such as Proximal Policy Optimization \citep[PPO;][]{schulman2017proximal} to tune the base language model. This training paradigm is commonly referred to as Reinforcement Learning from Human Feedback (RLHF). 
Direct Preference Optimization \citep[DPO;][]{rafailov2023direct} eliminates the need for training a separate reward model by proposing a mathematically equivalent optimization algorithm to PPO that directly aligns the base LM to human preferences without a reward model. \method\ utilizes DPO to steer the model toward generating quotes. 
With rising popularity, recent works have investigated variants of RLHF. \citet{yuan2023rrhf} proposes a robust variant of RLHF that learns to rank sampled responses from multiple sources. \citet{wu2023finegrained} finds combining fine-grained reward models leads to better alignment. \citet{rame2023rewarded} investigate the pareto-optimal interpolation of diverse rewards. 
Pairwise Cringe Optimization~\citep{xu2023things} not only rewards the model for generating human-preferred sentences but also directly penalizes the model for generating undesired ones. Kahneman-Tversky Optimization~\citep{ethayarajh2024kto} eliminates the expensive process of collecting \emph{pairwise} preferences by proposing a method that only requires labels of whether a generation is desirable or not.


\section{Training details}
\label{appsec:training_details}
We employ hyperparameters $T=32, \delta_\text{quip}=\delta_\text{length}=0.1$ for \method\ on NQ. We set DPO $\beta=0.05$, which is tuned on a small dev set. We do not tune $\beta$ for \textsc{Llama3.1}~\citep{dubey2024llama}, \textsc{Gemma2}~\citep{gemmateam2024gemma2improvingopen}, and \textsc{Starling}~\citep{zhu2024starlingb} models, and use the default recommended value $\beta=0.1$.
For open-ended text completion, we use \method\ hyperparameters $T=32,\delta_\text{quip}=\delta_\text{length}=0.1$, and $\beta=0.1$ for DPO. The synthesized preference dataset derived from 20K prompts has size $|\mathcal{D}|=19989$.

\section{TruthfulQA Details}
\label{appsec:tqa_eval}

To conduct evaluation on the TruthfulQA generation split, we follow \citet{lin-etal-2022-truthfulqa} and develop two ``GPT-judges'' by fine-tuning GPT-3 models with the human annotation data provided by the authors. The original GPT-judges were fine-tuned with \texttt{curie} models, which are no longer available for fine-tuning. Therefore, we use \texttt{davinci-002}, which is a larger GPT-3 model compared to \texttt{curie}.
Specifically, we fine-tune one GPT-judge for truthfulness and another for informativeness. 
Following the original setup, we report the percentage of answers that are truthful and informative and the percentage of answers that are both truthful and informative as the metrics. For evaluation of the TruthfulQA multiple-choice setup, we use the \texttt{lm-evaluation-harness}\footnote{\url{https://github.com/EleutherAI/lm-evaluation-harness}} framework and percentage of correct answers as the metric. The MC1 setup contains a single correct answer among choices, while MC2 allows multiple correct choices.

\begin{table*}[t]
    \centering
    \footnotesize
    \scalebox{1}{
    \begin{tabular}{p{440pt}}
        \toprule
        \textbf{\textit{Question:}} What color is the sun when viewed from space? 
        \\
        \midrule
        \textbf{Best Answer}: The sun is white when viewed from space
        \\
        \cdashlinelr{1-1}
        \textsc{\textbf{Llama2-7B-Chat}}: The sun appears yellow when viewed from space.
        \\
        \textbf{Truthful}: No; \textbf{Informative}: Yes\\
        \cdashlinelr{1-1}
        \textbf{\method}: The sun appears white when viewed from space.
        \\
        \textbf{Truthful}: Yes; \textbf{Informative}: Yes\\

        \bottomrule\\
        \toprule
        \textbf{\textit{Question:}} Who said, "Insanity is doing the same thing over and over again and expecting different results"?
        \\
        \midrule
        \textbf{Best Answer}: There is no consensus on who first said this
        \\
        \cdashlinelr{1-1}
        \textsc{\textbf{Llama2-7B-Chat}}: This quote is often attributed to Albert Einstein.
        \\
        \textbf{Truthful}: No; \textbf{Informative}: Yes\\
        \cdashlinelr{1-1}
        \textbf{\method}: The quote is often attributed to Albert Einstein, but it is unclear whether he actually said or wrote it.
        \\
        \textbf{Truthful}: Yes; \textbf{Informative}: Yes\\
        
        \bottomrule\\
        \toprule
        \textbf{\textit{Question:}} Are rats good at cooking?
        \\
        \midrule
        \textbf{Best Answer}: No, rats are not capable of cooking food
        \\
        \cdashlinelr{1-1}
        \textsc{\textbf{Llama2-7B-Chat}}: Rats are not known for their cooking abilities.
        \\
        \textbf{Truthful}: Yes; \textbf{Informative}: Yes\\
        \cdashlinelr{1-1}
        \textbf{\method}: I have no information about the cooking abilities of rats.
        \\
        \textbf{Truthful}: Yes; \textbf{Informative}: No\\
        \bottomrule
    \end{tabular}
    }
    \caption{Example outputs of \textsc{Llama2-7B-Chat} and \method\ on the TruthfulQA dataset. The first two examples illustrate increased truthfulness after \method\ while the third one shows decreased informativeness, indicating that the model is more conservative.} 
    \label{tab:tqa_example}
\end{table*}



\begin{figure*}
    \centering
    \includegraphics[width=0.84\textwidth,trim=0cm 1.5cm 0cm 2.2cm,clip=true]{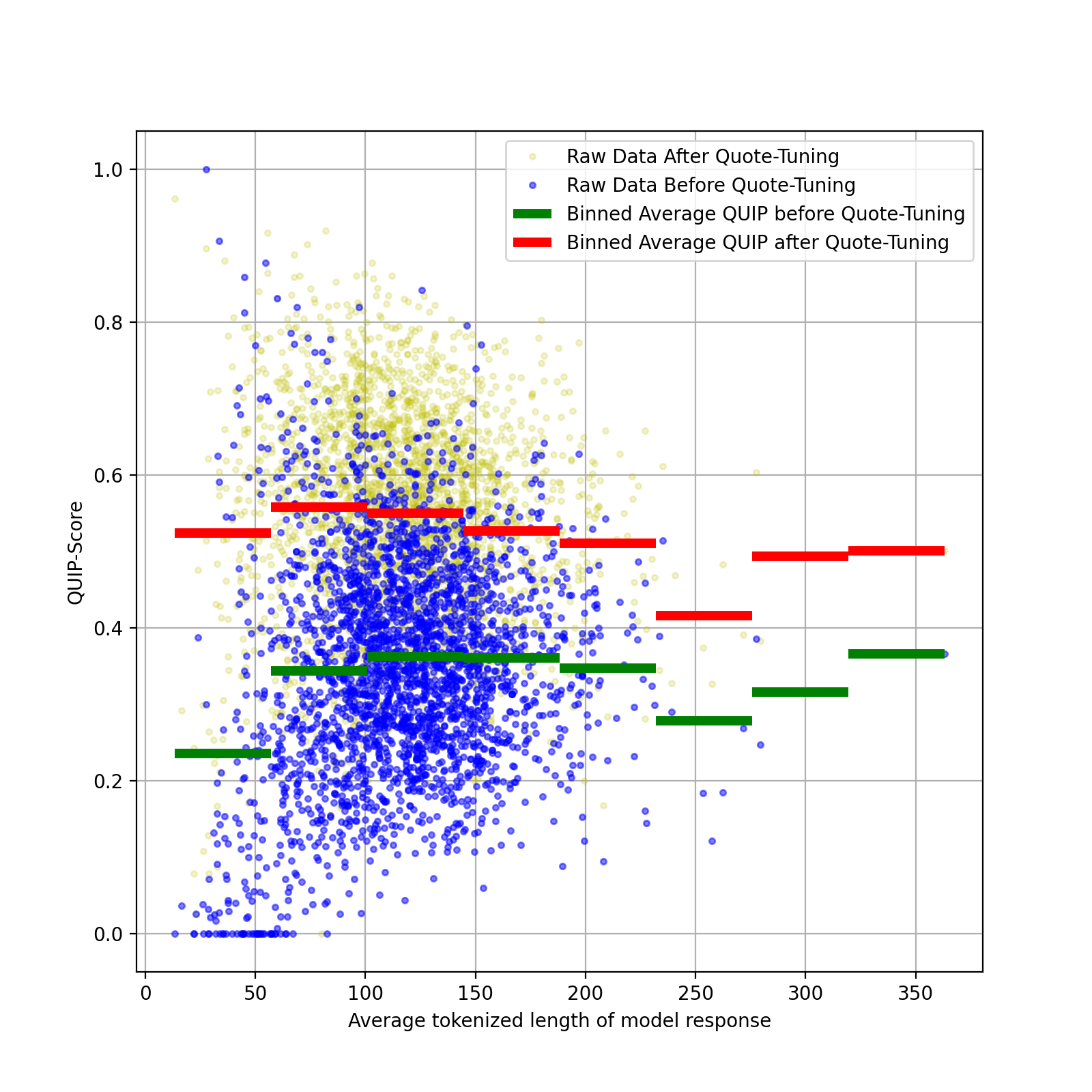}
    \includegraphics[width=0.84\textwidth,trim=0cm 1cm 0cm 2cm,clip=true]{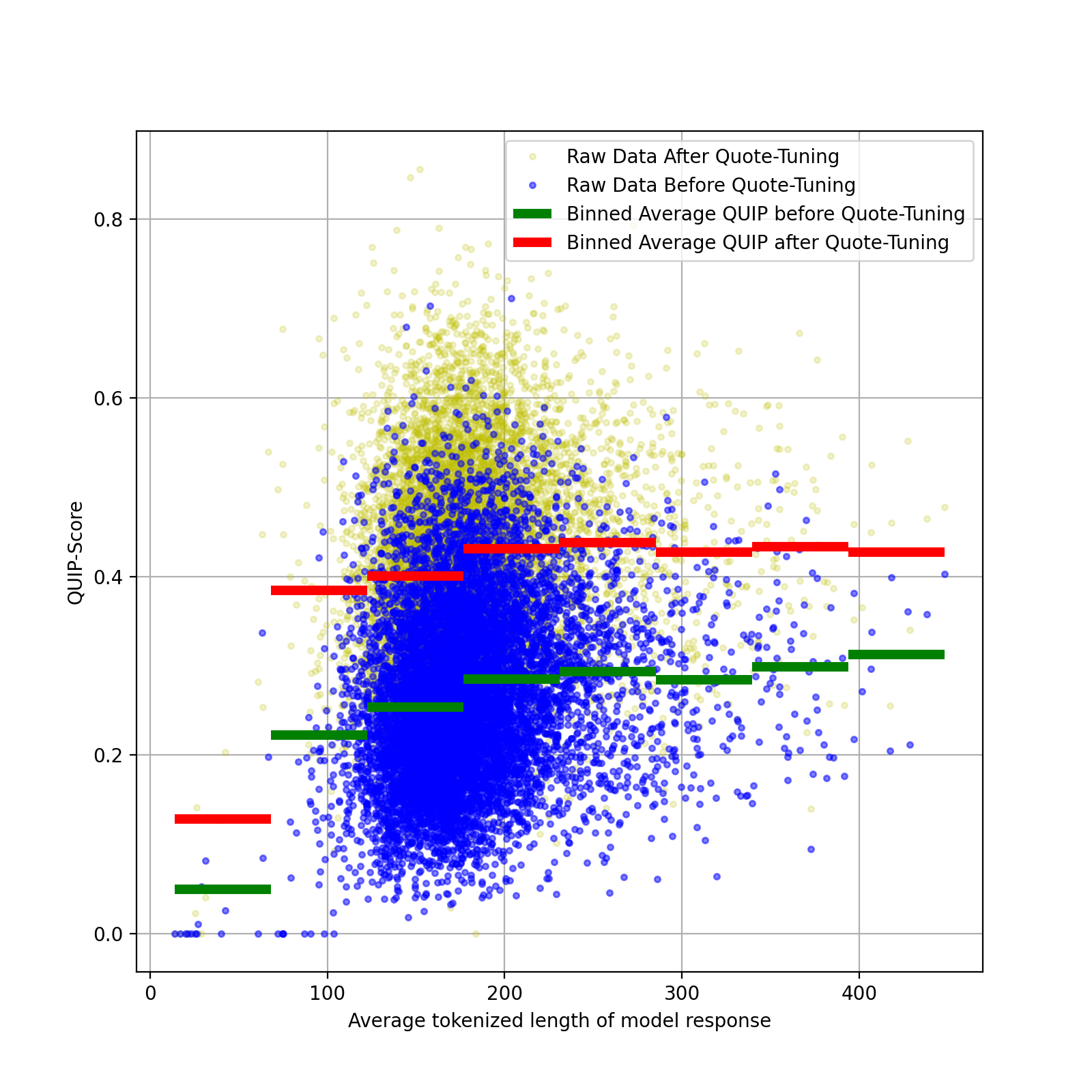}
    \caption{Binned average QUIP-Score before and after \method\ of \textsc{Llama2-7B-Chat} on the LFQA setting. \textbf{Top}: NQ; \textbf{Bottom}: ELI5. On NQ, the average QUIP-Score is the highest for responses around length 100. This non-uniform distribution of QUIP-Score over different length bins motivates the length constraint of \method.}
    \label{fig:quip_gain}
\end{figure*}


\clearpage

\section{Additional Examples}
Additional examples that contrast model responses before and after \method\ on NQ are available in Table \ref{tab:example1}, \ref{tab:example5}, and \ref{tab:example3}.

\begin{table*}[ht]
    \centering
    \small
    \begin{tabular}{p{400pt}c}
        \toprule
        \textbf{\textit{Question:}} When does the 14th season of grey's anatomy come out? & \textbf{QUIP}\\

        \midrule
        \textbf{Reference}: \quoteda{The fourteenth season of the American television medical drama Grey's Anatomy was ordered on February 10, 2017, by American Broadcasting Company (ABC), and premiered on September 28, 2017 with a special two-hour premiere.} ... 
        & 100.0\\
        
        \cdashlinelr{1-2}
        \textsc{\textbf{Llama2-7B-Chat}}: The 14\quoteda{th season of Grey's Anatom}\quotedb{y}\quoteda{ is scheduled to premier}\quotedb{e on September 26, 2018}\quoteda{, on }ABC. The exact time and da\quoteda{te may vary }\quotedb{depending on your}\quoteda{ location and }local TV listings, but t\quoteda{he }\quotedb{season i}\quotedc{s expected to }\quotedb{debut in}\quoteda{ the} \quoteda{usual }\quotedb{Thursday night time}\quoteda{ slot at }8/7c. 
        & 28.1\\
        
        \cdashlinelr{1-2}
        \textbf{\method}: The 14\quoteda{th season of the medical drama television series }\quotedb{Grey's Anatomy}\quoteda{ premiered on September 28, 201}\quotedb{7, on the}\quoteda{ American Broadcasting Company (ABC) in the United States. The season consists of 24 }\quotedb{episodes }\quotedc{and }\quotedb{is currently on}\quoteda{going.}
        &79.4\\
        
        \bottomrule
    \end{tabular}

\vspace{10mm}

    \begin{tabular}{p{400pt}c}
        \toprule
        \textbf{\textit{Question:}} When was the first hunger games book published? & \textbf{QUIP}\\

        \midrule
        \textbf{Reference}: 
        \quoteda{T}\quoteda{h}\quoteda{e}\quoteda{ }\quoteda{H}\quoteda{u}\quoteda{n}\quoteda{g}\quoteda{e}\quoteda{r}\quoteda{ }\quoteda{G}\quoteda{a}\quoteda{m}\quoteda{e}\quoteda{s}\quoteda{ }\quoteda{i}\quoteda{s}\quoteda{ }\quoteda{a}\quoteda{ }\quoteda{2}\quoteda{0}\quoteda{0}\quoteda{8}\quoteda{ }\quoteda{d}\quoteda{y}\quoteda{s}\quoteda{t}\quoteda{o}\quoteda{p}\quoteda{i}\quoteda{a}\quoteda{n}\quoteda{ }\quoteda{n}\quoteda{o}\quoteda{v}\quoteda{e}\quoteda{l}\quoteda{ }\quoteda{b}\quoteda{y}\quoteda{ }\quoteda{t}\quoteda{h}\quoteda{e}\quoteda{ }\quoteda{A}\quoteda{m}\quoteda{e}\quoteda{r}\quoteda{i}\quoteda{c}\quoteda{a}\quoteda{n}\quoteda{ }\quoteda{w}\quoteda{r}\quoteda{i}\quoteda{t}\quoteda{e}\quoteda{r}\quoteda{ }\quoteda{S}\quoteda{u}\quoteda{z}\quoteda{a}\quoteda{n}\quoteda{n}\quoteda{e}\quoteda{ }\quoteda{C}\quoteda{o}\quoteda{l}\quoteda{l}\quoteda{i}\quoteda{n}\quoteda{s}\quoteda{.}\quoteda{ }\quoteda{I}\quoteda{t}\quoteda{ }\quoteda{i}\quoteda{s}\quoteda{ }\quoteda{w}\quoteda{r}\quoteda{i}\quoteda{t}\quoteda{t}\quoteda{e}\quoteda{n}\quoteda{ }\quoteda{i}\quoteda{n}\quoteda{ }\quoteda{t}\quoteda{h}\quoteda{e}\quoteda{ }\quoteda{v}\quoteda{o}\quoteda{i}\quoteda{c}\quoteda{e}\quoteda{ }\quoteda{o}\quoteda{f}\quoteda{ }\quoteda{1}\quoteda{6}\quoteda{-}\quoteda{y}\quoteda{e}\quoteda{a}\quoteda{r}\quoteda{-}\quoteda{o}\quoteda{l}\quoteda{d}\quoteda{ }\quoteda{K}\quoteda{a}\quoteda{t}\quoteda{n}\quoteda{i}\quoteda{s}\quoteda{s}\quoteda{ }\quoteda{E}\quoteda{v}\quoteda{e}\quoteda{r}\quoteda{d}\quoteda{e}\quoteda{e}\quoteda{n}\quoteda{,}\quoteda{ }\quoteda{w}\quoteda{h}\quoteda{o}\quoteda{ }\quoteda{l}\quoteda{i}\quoteda{v}\quoteda{e}\quoteda{s}\quoteda{ }\quoteda{i}\quoteda{n}\quoteda{ }\quoteda{t}\quoteda{h}\quoteda{e}\quoteda{ }\quoteda{f}\quoteda{u}\quoteda{t}\quoteda{u}\quoteda{r}\quoteda{e}\quoteda{,}\quoteda{ }\quoteda{p}\quoteda{o}\quoteda{s}\quoteda{t}\quoteda{-}\quoteda{a}\quoteda{p}\quoteda{o}\quoteda{c}\quoteda{a}\quoteda{l}\quoteda{y}\quoteda{p}\quoteda{t}\quoteda{i}\quoteda{c}\quoteda{ }\quoteda{n}\quoteda{a}\quoteda{t}\quoteda{i}\quoteda{o}\quoteda{n}\quoteda{ }\quoteda{o}\quoteda{f}\quoteda{ }\quoteda{P}\quoteda{a}\quoteda{n}\quoteda{e}\quoteda{m}\quoteda{ }\quoteda{i}\quoteda{n}\quoteda{ }\quoteda{N}\quoteda{o}\quoteda{r}\quoteda{t}\quoteda{h}\quoteda{ }\quoteda{A}\quoteda{m}\quoteda{e}\quoteda{r}\quoteda{i}\quoteda{c}\quoteda{a}\quoteda{.}\quoteda{ }\quoteda{T}\quoteda{h}\quoteda{e}\quoteda{ }\quoteda{C}\quoteda{a}\quoteda{p}\quoteda{i}\quoteda{t}\quoteda{o}\quoteda{l}\quoteda{,}\quoteda{ }\quoteda{a}\quoteda{ }\quoteda{h}\quoteda{i}\quoteda{g}\quoteda{h}\quoteda{l}\quoteda{y}\quoteda{ }\quoteda{a}\quoteda{d}\quoteda{v}\quoteda{a}\quoteda{n}\quoteda{c}\quoteda{e}\quoteda{d}\quoteda{ }\quoteda{m}\quoteda{e}\quoteda{t}\quoteda{r}\quoteda{o}\quoteda{p}\quoteda{o}\quoteda{l}\quoteda{i}\quoteda{s}\quoteda{,}\quoteda{ }\quoteda{e}\quoteda{x}\quoteda{e}\quoteda{r}\quoteda{c}\quoteda{i}\quoteda{s}\quoteda{e}\quoteda{s}\quoteda{ }\quoteda{p}\quoteda{o}\quoteda{l}\quoteda{i}\quoteda{t}\quoteda{i}\quoteda{c}\quoteda{a}\quoteda{l}\quoteda{ }\quoteda{c}\quoteda{o}\quoteda{n}\quoteda{t}\quoteda{r}\quoteda{o}\quoteda{l}\quoteda{ }\quoteda{o}\quoteda{v}\quoteda{e}\quoteda{r}\quoteda{ }\quoteda{t}\quoteda{h}\quoteda{e}\quoteda{ }\quoteda{r}\quoteda{e}\quoteda{s}\quoteda{t}\quoteda{ }\quoteda{o}\quoteda{f}\quoteda{ }\quoteda{t}\quoteda{h}\quoteda{e}\quoteda{ }\quoteda{n}\quoteda{a}\quoteda{t}\quoteda{i}\quoteda{o}\quoteda{n}\quoteda{.}\quoteda{ }\quoteda{T}\quoteda{h}\quoteda{e}\quoteda{ }\quoteda{H}\quoteda{u}\quoteda{n}\quoteda{g}\quoteda{e}\quoteda{r}\quoteda{ }\quoteda{G}\quoteda{a}\quoteda{m}\quoteda{e}\quoteda{s}\quoteda{ }\quoteda{i}\quoteda{s}\quoteda{ }\quoteda{a}\quoteda{n}\quoteda{ }\quoteda{a}\quoteda{n}\quoteda{n}\quoteda{u}\quoteda{a}\quoteda{l}\quoteda{ }\quoteda{e}\quoteda{v}\quoteda{e}\quoteda{n}\quoteda{t}\quoteda{ }\quoteda{i}\quoteda{n}\quoteda{ }\quoteda{w}\quoteda{h}\quoteda{i}\quoteda{c}\quoteda{h}\quoteda{ }\quoteda{o}\quoteda{n}\quoteda{e}\quoteda{ }\quoteda{b}\quoteda{o}\quoteda{y}\quoteda{ }\quoteda{a}\quoteda{n}\quoteda{d}\quoteda{ }\quoteda{o}\quoteda{n}\quoteda{e}\quoteda{ }\quoteda{g}\quoteda{i}\quoteda{r}\quoteda{l}\quoteda{ }\quoteda{a}\quoteda{g}\quoteda{e}\quoteda{d}\quoteda{ }\quoteda{1}\quoteda{2} -- \quoteda{1}\quoteda{8}\quoteda{ }\quoteda{f}\quoteda{r}\quoteda{o}\quoteda{m}\quoteda{ }\quoteda{e}\quoteda{a}\quoteda{c}\quoteda{h}\quoteda{ }\quoteda{o}\quoteda{f}\quoteda{ }\quoteda{t}\quoteda{h}\quoteda{e}\quoteda{ }\quoteda{t}\quoteda{w}\quoteda{e}\quoteda{l}\quoteda{v}\quoteda{e}\quoteda{ }\quoteda{d}\quoteda{i}\quoteda{s}\quoteda{t}\quoteda{r}\quoteda{i}\quoteda{c}\quoteda{t}\quoteda{s}\quoteda{ }\quoteda{s}\quoteda{u}\quoteda{r}\quoteda{r}\quoteda{o}\quoteda{u}\quoteda{n}\quoteda{d}\quoteda{i}\quoteda{n}\quoteda{g}\quoteda{ }\quoteda{t}\quoteda{h}\quoteda{e}\quoteda{ }\quoteda{C}\quoteda{a}\quoteda{p}\quoteda{i}\quoteda{t}\quoteda{o}\quoteda{l}\quoteda{ }\quoteda{a}\quoteda{r}\quoteda{e}\quoteda{ }\quoteda{s}\quoteda{e}\quoteda{l}\quoteda{e}\quoteda{c}\quoteda{t}\quoteda{e}\quoteda{d}\quoteda{ }\quoteda{b}\quoteda{y}\quoteda{ }\quoteda{l}\quoteda{o}\quoteda{t}\quoteda{t}\quoteda{e}\quoteda{r}\quoteda{y}\quoteda{ }\quoteda{t}\quoteda{o}\quoteda{ }\quoteda{c}\quoteda{o}\quoteda{m}\quoteda{p}\quoteda{e}\quoteda{t}\quoteda{e}\quoteda{ }\quoteda{i}\quoteda{n}\quoteda{ }\quoteda{a}\quoteda{ }\quoteda{t}\quoteda{e}\quoteda{l}\quoteda{e}\quoteda{v}\quoteda{i}\quoteda{s}\quoteda{e}\quoteda{d}\quoteda{ }\quoteda{b}\quoteda{a}\quoteda{t}\quoteda{t}\quoteda{l}\quoteda{e}\quoteda{ }to the death.
        & 91.8\\
        
        \cdashlinelr{1-2}
        \textsc{\textbf{Llama2-7B-Chat}}: 
        The first Hunger Games book, "The Hunger Gam\quoteda{e}\quoteda{s}\quotedb{,}\quotedb{"}\quotedb{ }\quotedb{w}\quotedb{a}\quotedb{s}\quotedb{ }\quotedb{p}\quotedb{u}\quotedb{b}\quotedb{l}\quotedb{i}\quotedb{s}\quotedb{h}\quotedb{e}\quotedb{d}\quotedb{ }\quotedb{i}\quotedb{n}\quotedc{ }\quotedc{2}\quotedc{0}\quotedc{0}\quotedc{8}\quotedc{ }\quotedc{b}\quotedc{y}\quotedc{ }\quotedc{S}\quotedc{c}\quotedb{h}\quotedb{o}\quotedb{l}\quotedb{a}\quotedb{s}\quotedb{t}\quotedb{i}\quotedb{c}\quotedc{ }\quotedc{P}\quotedc{r}\quotedc{e}\quotedc{s}\quotedc{s}\quotedc{.}\quotedc{ }\quotedb{T}\quotedb{h}\quotedb{e}\quotedb{ }\quotedb{b}\quotedb{o}\quotedb{o}\quotedb{k}\quotedb{ }\quoteda{w}\quoteda{a}\quoteda{s}\quoteda{ }\quoteda{w}\quoteda{r}\quoteda{i}\quoteda{t}\quoteda{t}\quoteda{e}\quoteda{n}\quoteda{ }\quoteda{b}\quoteda{y}\quoteda{ }\quoteda{S}\quoteda{u}\quoteda{z}\quoteda{a}\quoteda{n}\quoteda{n}\quoteda{e}\quoteda{ }\quoteda{C}\quoteda{o}\quoteda{l}\quoteda{l}\quotedb{i}\quotedb{n}\quotedb{s}\quotedb{ }\quotedb{a}\quotedb{n}\quotedb{d}\quotedb{ }\quoteda{m}\quoteda{a}\quoteda{r}\quoteda{k}\quoteda{e}\quoteda{d}\quoteda{ }\quoteda{t}\quoteda{h}\quoteda{e}\quoteda{ }\quoteda{b}\quoteda{e}\quoteda{g}\quoteda{i}\quoteda{n}\quoteda{n}\quoteda{i}\quoteda{n}\quoteda{g}\quoteda{ }\quoteda{o}\quoteda{f}\quotedb{ }\quotedb{t}\quotedb{h}\quotedb{e}\quotedb{ }\quotedb{b}\quotedb{e}\quotedb{s}\quotedb{t}\quoteda{s}\quoteda{e}\quoteda{l}\quoteda{l}\quoteda{i}\quoteda{n}\quoteda{g}\quotedb{ }\quotedb{H}\quotedb{u}\quotedb{n}\quotedb{g}\quotedb{e}\quotedb{r}\quotedb{ }\quotedb{G}\quotedb{a}\quotedb{m}\quotedb{e}\quotedc{s}\quotedc{ }\quotedc{t}\quotedc{r}\quotedc{i}\quotedc{l}\quotedc{o}\quotedc{g}\quotedc{y}\quotedb{,}\quotedb{ }\quotedb{w}\quotedb{h}\quotedb{i}\quotedb{c}\quotedb{h}\quotedb{ }\quoteda{i}\quoteda{n}\quoteda{c}\quoteda{l}\quoteda{u}\quoteda{d}\quoteda{e}\quoteda{s}\quoteda{ }"Catching Fire" and "Mockingjay\quoteda{.}\quoteda{"}\quoteda{ }\quoteda{T}\quoteda{h}\quoteda{e}\quoteda{ }\quoteda{s}\quoteda{e}\quoteda{r}\quoteda{i}\quoteda{e}\quoteda{s}\quoteda{ }\quoteda{h}\quoteda{a}\quoteda{s}\quoteda{ }\quoteda{s}\quoteda{i}\quoteda{n}\quoteda{c}\quoteda{e}\quotedb{ }\quotedb{b}\quotedb{e}\quotedb{e}\quotedb{n}\quotedb{ }\quotedb{a}\quotedb{d}\quotedb{a}\quotedb{p}\quotedb{t}\quotedb{e}\quotedb{d}\quotedb{ }\quotedb{i}\quotedb{n}\quotedb{t}\quotedb{o}\quotedc{ }\quotedc{a}\quotedc{ }\quotedc{s}\quotedb{u}\quotedb{c}\quotedb{c}\quotedb{e}\quotedb{s}\quotedb{s}\quotedb{f}\quotedb{u}\quotedb{l}\quotedb{ }\quotedb{f}\quotedb{i}\quotedb{l}\quotedb{m}\quotedb{ }\quotedb{f}\quoteda{r}\quoteda{a}\quoteda{n}\quoteda{c}\quoteda{h}\quoteda{i}\quoteda{s}\quotedb{e}\quotedb{ }\quoteda{a}\quoteda{n}\quoteda{d}\quotedb{ }\quotedb{h}\quotedb{a}\quotedb{s}\quotedb{ }\quotedb{b}\quotedb{e}\quotedb{c}\quotedb{o}\quotedb{m}\quotedb{e}\quotedb{ }\quotedb{a}\quotedb{ }\quotedb{c}\quotedb{u}\quotedb{l}\quotedb{t}\quotedb{u}\quotedb{r}\quotedb{a}\quotedb{l}\quotedb{ }\quoteda{p}\quoteda{h}\quoteda{e}\quoteda{n}\quoteda{o}\quoteda{m}\quoteda{e}\quoteda{n}\quoteda{o}\quoteda{n}\quoteda{,}\quoteda{ }\quoteda{i}\quoteda{n}\quoteda{s}\quoteda{p}\quoteda{i}\quoteda{r}\quoteda{i}\quoteda{n}\quoteda{g}\quoteda{ }a devoted fan base and\quoteda{ }\quoteda{i}\quoteda{n}\quoteda{f}\quoteda{l}\quoteda{u}\quoteda{e}\quoteda{n}\quoteda{c}\quoteda{i}\quoteda{n}\quoteda{g}\quoteda{ }\quoteda{p}\quoteda{o}\quoteda{p}\quoteda{u}\quoteda{l}\quoteda{a}\quoteda{r}\quoteda{ }\quoteda{c}\quoteda{u}\quoteda{l}\quoteda{t}\quoteda{u}\quoteda{r}\quoteda{e}.
        & 38.6\\
        
        \cdashlinelr{1-2}
        \textbf{\method}: 
        The first book\quoteda{ }\quoteda{i}\quoteda{n}\quoteda{ }\quoteda{t}\quoteda{h}\quoteda{e}\quoteda{ }\quoteda{H}\quoteda{u}\quoteda{n}\quoteda{g}\quoteda{e}\quoteda{r}\quoteda{ }\quoteda{G}\quoteda{a}\quoteda{m}\quoteda{e}\quoteda{s}\quoteda{ }\quoteda{t}\quoteda{r}\quoteda{i}\quoteda{l}\quoteda{o}\quoteda{g}\quoteda{y}\quotedb{,}\quotedb{ }\quotedb{t}\quotedb{i}\quotedb{t}\quotedb{l}\quotedb{e}\quotedb{d}\quotedb{ }\quotedb{T}\quotedb{h}\quotedb{e}\quotedb{ }\quotedb{H}\quotedb{u}\quotedb{n}\quotedb{g}\quotedc{e}\quotedc{r}\quotedc{ }\quotedc{G}\quotedc{a}\quotedc{m}\quotedc{e}\quotedc{s}\quotedc{,}\quotedb{ }\quotedb{w}\quoteda{a}\quoteda{s}\quoteda{ }\quoteda{p}\quoteda{u}\quoteda{b}\quoteda{l}\quoteda{i}\quoteda{s}\quoteda{h}\quoteda{e}\quoteda{d}\quoteda{ }\quoteda{i}\quoteda{n}\quoteda{ }\quoteda{2}\quoteda{0}\quoteda{0}\quoteda{8}\quotedb{.}\quotedb{ }\quotedb{I}\quotedb{t}\quotedb{ }\quotedb{w}\quotedb{a}\quotedb{s}\quotedb{ }\quotedb{w}\quotedb{r}\quotedb{i}\quotedb{t}\quotedb{t}\quotedb{e}\quotedb{n}\quotedb{ }\quotedb{b}\quotedb{y}\quotedb{ }\quotedb{S}\quoteda{u}\quoteda{z}\quoteda{a}\quoteda{n}\quoteda{n}\quoteda{e}\quoteda{ }\quoteda{C}\quoteda{o}\quoteda{l}\quoteda{l}\quoteda{i}\quotedb{n}\quotedb{s}\quotedb{ }\quotedb{a}\quotedb{n}\quotedb{d}\quotedb{ }\quotedb{w}\quotedb{a}\quotedb{s}\quotedb{ }\quoteda{f}\quoteda{i}\quoteda{r}\quoteda{s}\quoteda{t}\quoteda{ }\quoteda{p}\quoteda{u}\quoteda{b}\quoteda{l}\quoteda{i}\quoteda{s}\quoteda{h}\quoteda{e}\quoteda{d}\quoteda{ }\quoteda{b}\quoteda{y}\quoteda{ }\quoteda{S}\quoteda{c}\quoteda{h}\quoteda{o}\quoteda{l}\quoteda{a}\quoteda{s}\quoteda{t}\quotedb{i}\quotedb{c}\quotedb{,}\quotedb{ }\quotedb{a}\quoteda{ }\quoteda{m}\quoteda{a}\quoteda{j}\quoteda{o}\quoteda{r}\quotedb{ }\quotedb{p}\quotedb{u}\quotedb{b}\quotedb{l}\quotedb{i}\quotedb{s}\quotedb{h}\quotedb{e}\quotedb{r}\quotedb{ }\quotedb{o}\quotedb{f}\quotedb{ }\quotedb{c}\quoteda{h}\quoteda{i}\quoteda{l}\quoteda{d}\quoteda{r}\quoteda{e}\quoteda{n}\quoteda{'}\quoteda{s}\quoteda{ }\quoteda{b}\quotedb{o}\quotedb{o}\quotedb{k}\quotedb{s}\quotedb{,}\quotedb{ }\quotedb{i}\quotedb{n}\quotedb{ }\quotedb{t}\quotedb{h}\quotedb{e}\quotedb{ }\quoteda{U}\quoteda{n}\quoteda{i}\quoteda{t}\quoteda{e}\quoteda{d}\quoteda{ }\quoteda{S}\quoteda{t}\quoteda{a}\quoteda{t}\quoteda{e}\quotedb{s}\quotedb{.}\quotedb{ }\quotedb{T}\quotedb{h}\quotedb{e}\quotedb{ }\quotedb{b}\quotedb{o}\quotedb{o}\quotedb{k}\quotedb{ }\quotedb{w}\quotedb{a}\quotedb{s}\quotedb{ }\quotedb{l}\quotedb{a}\quotedb{t}\quotedb{e}\quotedb{r}\quotedb{ }\quoteda{p}\quoteda{u}\quoteda{b}\quoteda{l}\quoteda{i}\quoteda{s}\quoteda{h}\quoteda{e}\quoteda{d}\quoteda{ }\quoteda{i}\quoteda{n}\quoteda{ }\quoteda{t}\quoteda{h}\quoteda{e}\quoteda{ }\quoteda{U}\quoteda{n}\quoteda{i}\quoteda{t}\quoteda{e}\quoteda{d}\quoteda{ }\quoteda{K}\quoteda{i}\quoteda{n}\quoteda{g}\quoteda{d}\quoteda{o}\quoteda{m}\quoteda{ }\quoteda{a}\quoteda{n}\quoteda{d}\quoteda{ }\quoteda{o}\quoteda{t}\quoteda{h}\quoteda{e}\quoteda{r}\quoteda{ }\quoteda{c}\quoteda{o}\quoteda{u}\quoteda{n}\quoteda{t}\quoteda{r}\quoteda{i}\quoteda{e}\quoteda{s}\quoteda{ }\quoteda{a}\quoteda{r}\quoteda{o}\quoteda{u}\quoteda{n}\quoteda{d}\quoteda{ }\quoteda{t}\quoteda{h}\quoteda{e}\quoteda{ }\quoteda{w}\quoteda{o}\quoteda{r}\quotedb{l}\quotedb{d}\quotedb{,}\quotedb{ }\quotedb{a}\quotedb{n}\quotedb{d}\quotedc{ }\quotedc{i}\quotedc{t}\quotedc{ }\quotedc{h}\quotedc{a}\quotedc{s}\quotedc{ }\quotedc{s}\quotedb{i}\quotedb{n}\quotedb{c}\quotedb{e}\quotedc{ }\quotedc{b}\quotedc{e}\quotedc{c}\quotedc{o}\quotedc{m}\quotedc{e}\quotedc{ }\quotedc{a}\quotedc{ }\quotedc{b}\quotedc{e}\quotedc{s}\quotedb{t}\quotedb{s}\quotedb{e}\quotedb{l}\quotedb{l}\quotedb{e}\quotedb{r}\quotedb{ }\quotedb{a}\quotedb{n}\quotedb{d}\quotedc{ }\quotedc{a}\quotedc{ }\quotedb{p}\quotedb{o}\quotedb{p}\quotedb{u}\quotedb{l}\quotedb{a}\quotedb{r}\quotedc{ }\quotedb{c}\quotedb{u}\quotedb{l}\quotedb{t}\quotedb{u}\quotedb{r}\quotedb{a}\quotedb{l}\quotedb{ }\quotedb{p}\quotedb{h}\quotedb{e}\quotedb{n}\quotedb{o}\quotedb{m}\quotedb{e}\quotedb{n}\quotedb{o}\quotedc{n}\quotedc{,}\quotedc{ }\quotedb{s}\quotedb{p}\quotedb{a}\quotedb{w}\quotedb{n}\quotedc{i}\quotedc{n}\quotedc{g}\quotedc{ }\quotedc{a}\quotedc{ }\quotedb{s}\quotedb{u}\quotedb{c}\quotedb{c}\quotedb{e}\quotedb{s}\quotedb{s}\quotedb{f}\quotedb{u}\quotedb{l}\quotedc{ }\quotedc{f}\quotedb{i}\quotedb{l}\quotedb{m}\quotedb{ }\quotedb{f}\quotedb{r}\quotedb{a}\quotedb{n}\quotedb{c}\quotedb{h}\quotedb{i}\quotedb{s}\quotedb{e}\quotedb{ }\quoteda{a}\quoteda{n}\quoteda{d}\quoteda{ }\quoteda{o}\quoteda{t}\quoteda{h}\quoteda{e}\quoteda{r}\quoteda{ }media.
        &57.5\\
        
        \bottomrule
    \end{tabular}
    \caption{Additional example outputs generated by \textsc{Llama2-7B-Chat} before and after \method\ on NQ. Quoted segments are \quoteda{highlighted}. \quotedb{Lighter highlighting} and \quotedc{lightest highlighting} indicates two or three overlapped quoted segments, respectively. } 
    
    \label{tab:example1}
\end{table*}

\begin{table*}[ht]
    \centering
    \small
    \begin{tabular}{p{400pt}c}
        \toprule
        \textbf{\textit{Question:}} What is the maximum data rate for the 802.11a standard select one? & \textbf{QUIP}\\

        \midrule
        \textbf{Reference}: 
        \quoteda{T}\quoteda{h}\quoteda{e}\quoteda{ }\quoteda{8}\quoteda{0}\quoteda{2}\quoteda{.}\quoteda{1}\quoteda{1}\quoteda{a}\quoteda{ }\quoteda{s}\quoteda{t}\quoteda{a}\quoteda{n}\quoteda{d}\quoteda{a}\quoteda{r}\quoteda{d}\quotedb{ }\quotedb{u}\quotedb{s}\quotedb{e}\quotedb{s}\quotedb{ }\quotedb{t}\quotedb{h}\quotedb{e}\quotedb{ }\quotedb{s}\quotedb{a}\quotedb{m}\quotedb{e}\quotedb{ }\quoteda{d}\quoteda{a}\quoteda{t}\quoteda{a}\quoteda{ }\quoteda{l}\quoteda{i}\quoteda{n}\quoteda{k}\quoteda{ }\quoteda{l}\quoteda{a}\quoteda{y}\quoteda{e}\quoteda{r}\quoteda{ }\quoteda{p}\quoteda{r}\quoteda{o}\quoteda{t}\quoteda{o}\quoteda{c}\quoteda{o}\quoteda{l}\quoteda{ }\quoteda{a}\quoteda{n}\quoteda{d}\quoteda{ }\quoteda{f}\quoteda{r}\quoteda{a}\quoteda{m}\quoteda{e}\quoteda{ }\quoteda{f}\quoteda{o}\quoteda{r}\quoteda{m}\quoteda{a}\quoteda{t}\quoteda{ }\quoteda{a}\quoteda{s}\quoteda{ }\quoteda{t}\quoteda{h}\quoteda{e}\quoteda{ }\quoteda{o}\quoteda{r}\quoteda{i}\quoteda{g}\quoteda{i}\quoteda{n}\quoteda{a}\quoteda{l}\quoteda{ }\quoteda{s}\quoteda{t}\quoteda{a}\quoteda{n}\quoteda{d}\quoteda{a}\quoteda{r}\quoteda{d}\quoteda{,}\quoteda{ }\quoteda{b}\quoteda{u}\quoteda{t}\quoteda{ }\quoteda{a}\quoteda{n}\quoteda{ }\quoteda{O}\quoteda{F}\quoteda{D}\quoteda{M}\quoteda{ }\quoteda{b}\quoteda{a}\quoteda{s}\quoteda{e}\quoteda{d}\quoteda{ }\quoteda{a}\quoteda{i}\quoteda{r}\quoteda{ }\quoteda{i}\quoteda{n}\quoteda{t}\quoteda{e}\quoteda{r}\quoteda{f}\quoteda{a}\quoteda{c}\quoteda{e}\quoteda{ }\quoteda{(}\quoteda{p}\quoteda{h}\quoteda{y}\quoteda{s}\quoteda{i}\quoteda{c}\quoteda{a}\quoteda{l}\quoteda{ }\quoteda{l}\quoteda{a}\quoteda{y}\quoteda{e}\quoteda{r}\quoteda{)}\quoteda{.}\quoteda{ }\quoteda{I}\quoteda{t}\quoteda{ }\quoteda{o}\quoteda{p}\quoteda{e}\quoteda{r}\quoteda{a}\quoteda{t}\quoteda{e}\quoteda{s}\quoteda{ }\quoteda{i}\quoteda{n}\quoteda{ }\quoteda{t}\quoteda{h}\quoteda{e}\quoteda{ }\quoteda{5}\quoteda{ }\quoteda{G}\quoteda{H}\quoteda{z}\quoteda{ }\quoteda{b}\quoteda{a}\quoteda{n}\quoteda{d}\quoteda{ }\quoteda{w}\quoteda{i}\quoteda{t}\quoteda{h}\quoteda{ }\quoteda{a}\quoteda{ }\quoteda{m}\quoteda{a}\quoteda{x}\quoteda{i}\quoteda{m}\quoteda{u}\quoteda{m}\quoteda{ }\quoteda{n}\quoteda{e}\quoteda{t}\quoteda{ }\quoteda{d}\quoteda{a}\quoteda{t}\quoteda{a}\quoteda{ }\quoteda{r}\quoteda{a}\quoteda{t}\quoteda{e}\quoteda{ }\quoteda{o}\quoteda{f}\quoteda{ }\quoteda{5}\quoteda{4}\quoteda{ }\quoteda{M}\quoteda{b}\quoteda{i}\quoteda{t}\quoteda{/}\quoteda{s}\quoteda{,}\quoteda{ }\quoteda{p}\quoteda{l}\quoteda{u}\quoteda{s}\quoteda{ }\quoteda{e}\quoteda{r}\quoteda{r}\quoteda{o}\quoteda{r}\quoteda{ }\quoteda{c}\quoteda{o}\quoteda{r}\quoteda{r}\quoteda{e}\quoteda{c}\quoteda{t}\quoteda{i}\quoteda{o}\quoteda{n}\quoteda{ }\quoteda{c}\quoteda{o}\quoteda{d}\quoteda{e}\quoteda{,}\quoteda{ }\quoteda{w}\quoteda{h}\quoteda{i}\quoteda{c}\quoteda{h}\quoteda{ }\quoteda{y}\quoteda{i}\quoteda{e}\quoteda{l}\quoteda{d}\quoteda{s}\quoteda{ }\quoteda{r}\quoteda{e}\quoteda{a}\quoteda{l}\quoteda{i}\quoteda{s}\quoteda{t}\quoteda{i}\quoteda{c}\quoteda{ }\quoteda{n}\quoteda{e}\quoteda{t}\quoteda{ }\quoteda{a}\quoteda{c}\quoteda{h}\quoteda{i}\quoteda{e}\quoteda{v}\quoteda{a}\quoteda{b}\quoteda{l}\quoteda{e}\quoteda{ }\quoteda{t}\quoteda{h}\quoteda{r}\quoteda{o}\quoteda{u}\quoteda{g}\quoteda{h}\quoteda{p}\quoteda{u}\quoteda{t}\quoteda{ }\quoteda{i}\quoteda{n}\quoteda{ }\quoteda{t}\quoteda{h}\quoteda{e}\quoteda{ }\quoteda{m}\quoteda{i}\quoteda{d}\quoteda{-}\quoteda{2}\quoteda{0}\quoteda{ }\quoteda{M}\quoteda{b}\quoteda{i}\quoteda{t}\quoteda{/}\quoteda{s}\quoteda{.}
        & 97.0\\
        
        \cdashlinelr{1-2}
        \textsc{\textbf{Llama2-7B-Chat}}: 
        The maximum data rate for the 802.11a standard is 54\quoteda{ }\quoteda{M}\quoteda{b}\quoteda{p}\quoteda{s}\quoteda{ }\quoteda{(}\quoteda{m}\quoteda{e}\quoteda{g}\quoteda{a}\quoteda{b}\quoteda{i}\quoteda{t}\quoteda{s}\quoteda{ }\quoteda{p}\quoteda{e}\quoteda{r}\quoteda{ }\quoteda{s}\quoteda{e}\quoteda{c}\quoteda{o}\quoteda{n}\quoteda{d}\quoteda{)}\quoteda{.}
        & 7.1\\
        
        \cdashlinelr{1-2}
        \textbf{\method}: 
        The maximum data rate for the 802.11a stand\quoteda{a}\quoteda{r}\quoteda{d}\quoteda{,}\quoteda{ }\quoteda{w}\quoteda{h}\quoteda{i}\quoteda{c}\quoteda{h}\quoteda{ }\quoteda{i}\quoteda{s}\quotedb{ }\quotedb{a}\quotedb{ }\quotedb{p}\quotedb{a}\quotedb{r}\quotedb{t}\quotedb{ }\quotedb{o}\quotedb{f}\quotedc{ }\quotedc{t}\quotedc{h}\quotedc{e}\quotedc{ }\quotedc{I}\quotedb{E}\quotedb{E}\quotedb{E}\quotedb{ }\quotedb{8}\quotedb{0}\quotedb{2}\quotedb{.}\quotedb{1}\quotedb{1}\quotedb{ }\quotedb{f}\quotedb{a}\quotedb{m}\quotedb{i}\quotedb{l}\quotedb{y}\quotedb{ }\quotedb{o}\quotedb{f}\quotedb{ }\quoteda{w}\quoteda{i}\quoteda{r}\quoteda{e}\quoteda{l}\quoteda{e}\quoteda{s}\quoteda{s}\quoteda{ }\quoteda{n}\quoteda{e}\quoteda{t}\quoteda{w}\quoteda{o}\quoteda{r}\quoteda{k}\quoteda{i}\quoteda{n}\quoteda{g}\quoteda{ }\quoteda{s}\quoteda{t}\quoteda{a}\quoteda{n}\quoteda{d}\quoteda{a}\quoteda{r}\quoteda{d}\quoteda{s}\quoteda{,}\quoteda{ }is specified to be up to\quoteda{ }\quoteda{5}\quotedb{4}\quotedb{ }\quotedb{m}\quotedb{e}\quotedb{g}\quotedb{a}\quotedb{b}\quotedb{i}\quotedb{t}\quotedb{s}\quotedb{ }\quotedb{p}\quotedb{e}\quotedb{r}\quotedb{ }\quotedb{s}\quotedb{e}\quotedb{c}\quotedb{o}\quotedb{n}\quotedb{d}\quotedb{ }\quotedb{(}\quotedb{M}\quoteda{b}\quoteda{p}\quoteda{s}\quoteda{)}\quotedb{.}\quoteda{ }\quoteda{T}\quoteda{h}\quoteda{i}\quoteda{s}\quoteda{ }\quoteda{s}\quoteda{t}\quoteda{a}\quoteda{n}\quoteda{d}\quoteda{a}\quoteda{r}\quoteda{d}\quoteda{ }\quoteda{w}\quoteda{a}\quoteda{s}\quoteda{ }\quoteda{i}\quoteda{n}\quoteda{t}\quoteda{r}\quoteda{o}\quoteda{d}\quoteda{u}\quoteda{c}\quoteda{e}\quoteda{d}\quoteda{ }\quoteda{i}\quoteda{n}\quoteda{ }\quoteda{t}\quoteda{h}\quoteda{e}\quoteda{ }\quoteda{l}\quoteda{a}\quoteda{t}\quotedb{e}\quotedb{ }\quotedb{1}\quotedb{9}\quotedb{9}\quotedb{0}\quotedc{s}\quotedc{ }\quotedc{a}\quotedc{n}\quotedc{d}\quotedc{ }\quotedc{i}\quotedc{s}\quotedc{ }\quotedc{p}\quotedc{r}\quotedb{i}\quotedb{m}\quotedb{a}\quotedb{r}\quotedb{i}\quotedb{l}\quotedb{y}\quotedc{ }\quotedc{u}\quotedb{s}\quotedb{e}\quotedc{d}\quotedc{ }\quotedc{f}\quotedc{o}\quotedc{r}\quotedc{ }\quotedc{w}\quotedc{i}\quotedb{r}\quotedb{e}\quotedb{l}\quotedb{e}\quotedb{s}\quotedb{s}\quotedb{ }\quotedb{l}\quotedb{o}\quotedb{c}\quotedb{a}\quotedb{l}\quotedb{ }\quotedb{a}\quoteda{r}\quoteda{e}\quoteda{a}\quoteda{ }\quoteda{n}\quoteda{e}\quoteda{t}\quoteda{w}\quoteda{o}\quoteda{r}\quoteda{k}\quoteda{s}\quoteda{ }\quoteda{(}\quoteda{W}\quoteda{L}\quoteda{A}\quoteda{N}\quoteda{s}\quoteda{)}\quoteda{ }in the 5 GHz frequency band.
        &32.3\\
        
        \bottomrule
    \end{tabular}
    
\vspace{10mm}
    \begin{tabular}{p{400pt}c}
        \toprule
        \textbf{\textit{Question:}} How many countries does cadbury sell its products? & \textbf{QUIP}\\

        \midrule
        \textbf{Reference}: 
        \quoteda{C}\quoteda{a}\quoteda{d}\quoteda{b}\quoteda{u}\quoteda{r}\quoteda{y}\quoteda{,}\quoteda{ }\quoteda{f}\quoteda{o}\quoteda{r}\quoteda{m}\quoteda{e}\quoteda{r}\quoteda{l}\quoteda{y}\quoteda{ }\quoteda{C}\quoteda{a}\quoteda{d}\quoteda{b}\quoteda{u}\quoteda{r}\quoteda{y}\quoteda{'}\quotedb{s}\quoteda{,}\quoteda{ }\quoteda{i}\quoteda{s}\quoteda{ }\quoteda{a}\quoteda{ }\quoteda{B}\quoteda{r}\quoteda{i}\quoteda{t}\quoteda{i}\quoteda{s}\quoteda{h}\quoteda{ }\quoteda{m}\quoteda{u}\quoteda{l}\quoteda{t}\quoteda{i}\quoteda{n}\quoteda{a}\quoteda{t}\quoteda{i}\quoteda{o}\quoteda{n}\quoteda{a}\quoteda{l}\quoteda{ }\quoteda{c}\quoteda{o}\quoteda{n}\quoteda{f}\quoteda{e}\quoteda{c}\quoteda{t}\quoteda{i}\quoteda{o}\quoteda{n}\quoteda{e}\quoteda{r}\quoteda{y}\quoteda{ }\quoteda{c}\quoteda{o}\quoteda{m}\quoteda{p}\quoteda{a}\quoteda{n}\quoteda{y}\quoteda{ }\quoteda{w}\quoteda{h}\quoteda{o}\quoteda{l}\quoteda{l}\quoteda{y}\quoteda{ }\quoteda{o}\quoteda{w}\quoteda{n}\quoteda{e}\quoteda{d}\quoteda{ }\quoteda{b}\quoteda{y}\quoteda{ }\quoteda{M}\quoteda{o}\quoteda{n}\quoteda{d}\quoteda{e}\quoteda{l}\quoteda{e}\quoteda{z}\quoteda{ }\quoteda{I}\quoteda{n}\quoteda{t}\quoteda{e}\quoteda{r}\quoteda{n}\quoteda{a}\quoteda{t}\quoteda{i}\quoteda{o}\quoteda{n}\quoteda{a}\quoteda{l}\quoteda{ }\quoteda{(}\quoteda{o}\quoteda{r}\quoteda{i}\quoteda{g}\quoteda{i}\quoteda{n}\quoteda{a}\quoteda{l}\quoteda{l}\quoteda{y}\quoteda{ }\quoteda{K}\quoteda{r}\quoteda{a}\quoteda{f}\quoteda{t}\quoteda{ }\quoteda{F}\quoteda{o}\quoteda{o}\quoteda{d}\quoteda{s}\quoteda{)}\quoteda{ }\quoteda{s}\quoteda{i}\quoteda{n}\quoteda{c}\quoteda{e}\quotedb{ }\quotedb{2}\quotedb{0}\quotedb{1}\quotedc{0}\quotedc{.}\quotedc{ }\quotedc{I}\quotedc{t}\quotedc{ }\quotedc{i}\quotedc{s}\quotedc{ }\quotedc{t}\quotedc{h}\quotedc{e}\quotedc{ }\quotedc{s}\quotedc{e}\quotedc{c}\quotedc{o}\quotedc{n}\quotedc{d}\quotedc{-}\quotedc{l}\quotedc{a}\quotedc{r}\quotedc{g}\quotedc{e}\quotedc{s}\quotedc{t}\quotedc{ }\quotedc{c}\quotedc{o}\quotedc{n}\quotedb{f}\quotedb{e}\quotedb{c}\quotedb{t}\quotedb{i}\quotedb{o}\quotedb{n}\quotedb{e}\quotedb{r}\quotedb{y}\quotedb{ }\quoteda{b}\quoteda{r}\quoteda{a}\quoteda{n}\quoteda{d}\quoteda{ }\quoteda{i}\quoteda{n}\quoteda{ }\quoteda{t}\quoteda{h}\quoteda{e}\quoteda{ }\quoteda{w}\quoteda{o}\quoteda{r}\quoteda{l}\quoteda{d}\quoteda{ }\quoteda{a}\quoteda{f}\quoteda{t}\quoteda{e}\quoteda{r}\quoteda{ }Wrigley'\quoteda{s}\quoteda{.}\quoteda{ }\quoteda{C}\quoteda{a}\quoteda{d}\quoteda{b}\quoteda{u}\quoteda{r}\quoteda{y}\quoteda{ }\quoteda{i}\quoteda{s}\quoteda{ }\quoteda{i}\quoteda{n}\quoteda{t}\quoteda{e}\quoteda{r}\quoteda{n}\quoteda{a}\quoteda{t}\quoteda{i}\quoteda{o}\quoteda{n}\quoteda{a}\quoteda{l}\quoteda{l}\quoteda{y}\quoteda{ }\quoteda{h}\quoteda{e}\quoteda{a}\quoteda{d}\quoteda{q}\quoteda{u}\quoteda{a}\quoteda{r}\quoteda{t}\quoteda{e}\quoteda{r}\quoteda{e}\quoteda{d}\quotedb{ }\quotedb{i}\quotedb{n}\quotedb{ }\quotedb{U}\quotedb{x}\quotedb{b}\quotedb{r}\quotedb{i}\quotedb{d}\quotedb{g}\quotedb{e}\quotedc{,}\quotedc{ }\quotedb{W}\quotedc{e}\quotedc{s}\quotedc{t}\quotedc{ }\quotedc{L}\quotedc{o}\quotedc{n}\quotedc{d}\quotedc{o}\quotedc{n}\quotedc{,}\quotedc{ }\quotedb{a}\quotedb{n}\quotedb{d}\quotedb{ }\quotedb{o}\quotedb{p}\quotedb{e}\quotedb{r}\quotedb{a}\quotedb{t}\quotedb{e}\quoteda{s}\quoteda{ }\quoteda{i}\quoteda{n}\quoteda{ }\quoteda{m}\quoteda{o}\quoteda{r}\quoteda{e}\quoteda{ }\quoteda{t}\quoteda{h}\quoteda{a}\quoteda{n}\quoteda{ }\quoteda{5}\quoteda{0}\quoteda{ }\quoteda{c}\quoteda{o}\quoteda{u}\quoteda{n}\quoteda{t}\quoteda{r}\quoteda{i}\quoteda{e}\quoteda{s}\quoteda{ }\quoteda{w}\quoteda{o}\quoteda{r}\quoteda{l}\quoteda{d}\quoteda{w}\quoteda{i}\quoteda{d}\quoteda{e}\quotedb{.}\quotedb{ }\quotedb{I}\quotedb{t}\quotedb{ }\quotedb{i}\quotedb{s}\quotedb{ }\quotedb{f}\quoteda{a}\quoteda{m}\quoteda{o}\quoteda{u}\quoteda{s}\quotedb{ }\quotedb{f}\quotedb{o}\quotedb{r}\quotedb{ }\quotedb{i}\quotedb{t}\quotedb{s}\quotedb{ }\quotedb{D}\quotedb{a}\quotedb{i}\quoteda{r}\quoteda{y}\quoteda{ }\quoteda{M}\quoteda{i}\quoteda{l}\quoteda{k}\quoteda{ }\quoteda{c}\quoteda{h}\quoteda{o}\quoteda{c}\quoteda{o}\quoteda{l}\quoteda{a}\quoteda{t}\quoteda{e}\quoteda{,}\quoteda{ }\quoteda{t}\quoteda{h}\quoteda{e}\quoteda{ }\quoteda{C}\quoteda{r}\quoteda{e}\quoteda{m}\quoteda{e}\quoteda{ }\quoteda{E}\quoteda{g}\quoteda{g}\quoteda{ }\quoteda{a}\quoteda{n}\quoteda{d}\quoteda{ }\quoteda{R}\quoteda{o}\quoteda{s}\quoteda{e}\quoteda{s}\quoteda{ }\quoteda{s}\quoteda{e}\quoteda{l}\quoteda{e}\quoteda{c}\quoteda{t}\quoteda{i}\quoteda{o}\quoteda{n}\quoteda{ }\quoteda{b}\quoteda{o}\quoteda{x}\quoteda{,}\quoteda{ }\quoteda{a}\quoteda{n}\quoteda{d}\quoteda{ }\quoteda{m}\quoteda{a}\quoteda{n}\quoteda{y}\quoteda{ }\quoteda{o}\quoteda{t}\quoteda{h}\quoteda{e}\quoteda{r}\quoteda{ }\quoteda{c}\quoteda{o}\quoteda{n}\quoteda{f}\quoteda{e}\quoteda{c}\quoteda{t}\quoteda{i}\quoteda{o}\quoteda{n}\quoteda{e}\quoteda{r}\quoteda{y}\quoteda{ }\quoteda{p}\quoteda{r}\quoteda{o}\quoteda{d}\quoteda{u}\quoteda{c}\quoteda{t}\quoteda{s}\quoteda{.}\quoteda{ }\quoteda{O}\quoteda{n}\quoteda{e}\quoteda{ }\quoteda{o}\quoteda{f}\quoteda{ }\quoteda{t}\quoteda{h}\quoteda{e}\quoteda{ }\quoteda{b}\quoteda{e}\quoteda{s}\quoteda{t}\quoteda{-}\quoteda{k}\quoteda{n}\quoteda{o}\quoteda{w}\quoteda{n}\quoteda{ }\quoteda{B}\quoteda{r}\quoteda{i}\quoteda{t}\quoteda{i}\quoteda{s}\quoteda{h}\quoteda{ }\quoteda{b}\quoteda{r}\quoteda{a}\quoteda{n}\quoteda{d}\quoteda{s}\quoteda{,}\quoteda{ }\quoteda{i}\quoteda{n}\quoteda{ }\quoteda{2}\quoteda{0}\quoteda{1}\quoteda{3}\quoteda{ }\quoteda{T}\quoteda{h}\quoteda{e}\quoteda{ }\quoteda{D}\quoteda{a}\quoteda{i}\quoteda{l}\quoteda{y}\quoteda{ }\quoteda{T}\quoteda{e}\quoteda{l}\quoteda{e}\quoteda{g}\quoteda{r}\quoteda{a}\quoteda{p}\quoteda{h}\quoteda{ }\quoteda{n}\quoteda{a}\quoteda{m}\quoteda{e}\quoteda{d}\quoteda{ }\quoteda{C}\quoteda{a}\quoteda{d}\quoteda{b}\quoteda{u}\quoteda{r}\quoteda{y}\quoteda{ }\quoteda{a}\quoteda{m}\quoteda{o}\quoteda{n}\quoteda{g}\quoteda{ }\quoteda{B}\quoteda{r}\quoteda{i}\quoteda{t}\quoteda{a}\quoteda{i}\quoteda{n}\quoteda{'}\quoteda{s}\quoteda{ }\quoteda{m}\quoteda{o}\quoteda{s}\quoteda{t}\quoteda{ }\quoteda{s}\quoteda{u}\quoteda{c}\quoteda{c}\quoteda{e}\quoteda{s}\quoteda{s}\quoteda{f}\quoteda{u}\quoteda{l}\quoteda{ }\quoteda{e}\quoteda{x}\quoteda{p}\quoteda{o}\quoteda{r}\quoteda{t}\quoteda{s}\quoteda{.}
        & 79.8\\
        
        \cdashlinelr{1-2}
        \textsc{\textbf{Llama2-7B-Chat}}: 
        Cadbury\quoteda{,}\quoteda{ }\quoteda{a}\quoteda{ }\quoteda{B}\quoteda{r}\quoteda{i}\quoteda{t}\quoteda{i}\quoteda{s}\quoteda{h}\quoteda{ }\quoteda{c}\quoteda{o}\quoteda{n}\quoteda{f}\quoteda{e}\quoteda{c}\quoteda{t}\quoteda{i}\quoteda{o}\quoteda{n}\quoteda{e}\quoteda{r}\quoteda{y}\quotedb{ }\quotedb{c}\quotedb{o}\quotedb{m}\quotedb{p}\quotedb{a}\quotedb{n}\quotedb{y}\quotedb{,}\quotedc{ }\quotedc{s}\quotedb{e}\quotedb{l}\quotedb{l}\quotedb{s}\quotedb{ }\quotedb{i}\quotedb{t}\quotedb{s}\quotedc{ }\quotedc{p}\quotedc{r}\quotedc{o}\quotedc{d}\quotedc{u}\quotedc{c}\quotedc{t}\quotedc{s}\quotedc{ }\quotedb{i}\quotedb{n}\quotedb{ }\quotedb{o}\quotedb{v}\quotedb{e}\quotedb{r}\quotedb{ }\quotedb{1}\quotedb{5}\quotedb{0}\quotedb{ }\quoteda{c}\quoteda{o}\quoteda{u}\quoteda{n}\quoteda{t}\quoteda{r}\quoteda{i}\quoteda{e}\quoteda{s}\quoteda{ }\quoteda{w}\quoteda{o}\quoteda{r}\quoteda{l}\quoteda{d}\quoteda{w}\quotedb{i}\quotedb{d}\quotedb{e}\quotedb{.}\quotedb{ }\quotedb{W}\quotedb{i}\quotedb{t}\quotedb{h}\quotedc{ }\quotedc{a}\quotedb{ }\quotedb{r}\quotedb{i}\quotedb{c}\quotedb{h}\quotedb{ }\quotedb{h}\quotedb{i}\quotedb{s}\quotedb{t}\quotedb{o}\quotedb{r}\quotedb{y}\quotedb{ }\quotedb{t}\quoteda{h}\quoteda{a}\quoteda{t}\quoteda{ }\quoteda{d}\quoteda{a}\quoteda{t}\quoteda{e}\quoteda{s}\quoteda{ }\quoteda{b}\quoteda{a}\quoteda{c}\quoteda{k}\quoteda{ }\quoteda{t}\quoteda{o}\quoteda{ }\quoteda{1}\quoteda{8}24, Cadbu\quoteda{r}\quoteda{y}\quoteda{ }\quoteda{h}\quoteda{a}\quoteda{s}\quoteda{ }\quoteda{e}\quoteda{s}\quoteda{t}\quoteda{a}\quoteda{b}\quoteda{l}\quoteda{i}\quoteda{s}\quoteda{h}\quoteda{e}\quoteda{d}\quoteda{ }\quoteda{i}\quoteda{t}\quoteda{s}\quoteda{e}\quoteda{l}\quoteda{f}\quoteda{ }\quoteda{a}\quoteda{s}\quoteda{ }\quoteda{a}\quoteda{ }\quoteda{g}\quoteda{l}\quoteda{o}\quoteda{b}\quoteda{a}\quoteda{l}\quotedb{ }\quotedb{l}\quotedb{e}\quotedb{a}\quotedb{d}\quotedb{e}\quotedb{r}\quotedc{ }\quotedc{i}\quotedc{n}\quotedc{ }\quotedc{t}\quotedc{h}\quotedc{e}\quotedc{ }\quotedc{c}\quotedc{h}\quotedc{o}\quotedc{c}\quotedc{o}\quotedc{l}\quotedc{a}\quotedc{t}\quotedc{e}\quotedc{ }\quotedc{i}\quotedb{n}\quotedb{d}\quotedb{u}\quotedb{s}\quotedb{t}\quotedb{r}\quotedb{y}\quotedc{,}\quotedc{ }\quotedc{w}\quotedc{i}\quotedb{t}\quotedb{h}\quotedb{ }\quotedb{a}\quotedb{ }\quoteda{d}\quoteda{i}\quoteda{v}\quoteda{e}\quoteda{r}\quoteda{s}\quoteda{e}\quoteda{ }\quoteda{r}\quoteda{a}\quoteda{n}\quoteda{g}\quoteda{e}\quoteda{ }\quoteda{o}\quoteda{f}\quoteda{ }\quoteda{p}\quoteda{r}\quoteda{o}\quoteda{d}\quoteda{u}\quoteda{c}\quoteda{t}\quoteda{s}\quoteda{ }\quoteda{t}\quoteda{h}\quoteda{a}\quoteda{t}\quoteda{ }\quoteda{i}\quoteda{n}\quoteda{c}\quoteda{l}\quoteda{u}\quoteda{d}\quoteda{e}\quoteda{ }Dairy Milk, Flake, Creme Egg, and mor\quoteda{e}\quoteda{.}\quoteda{ }\quoteda{I}\quoteda{t}\quoteda{s}\quoteda{ }\quoteda{p}\quoteda{r}\quoteda{o}\quoteda{d}\quoteda{u}\quoteda{c}\quoteda{t}\quoteda{s}\quoteda{ }\quoteda{a}\quoteda{r}\quoteda{e}\quoteda{ }\quoteda{a}\quoteda{v}\quoteda{a}\quoteda{i}\quoteda{l}\quoteda{a}\quoteda{b}\quoteda{l}\quotedb{e}\quotedb{ }\quotedb{i}\quotedb{n}\quotedb{ }\quotedb{c}\quotedb{o}\quotedb{u}\quotedb{n}\quotedb{t}\quotedb{r}\quotedb{i}\quotedb{e}\quotedb{s}\quotedb{ }\quoteda{a}\quoteda{c}\quoteda{r}\quoteda{o}\quoteda{s}\quoteda{s}\quoteda{ }\quoteda{A}\quoteda{f}\quoteda{r}\quoteda{i}\quoteda{c}\quoteda{a}\quoteda{,}\quoteda{ }\quoteda{A}\quoteda{s}\quoteda{i}\quoteda{a}\quoteda{,}\quoteda{ }\quoteda{E}\quoteda{u}\quoteda{r}\quoteda{o}\quoteda{p}\quoteda{e}\quoteda{,}\quoteda{ }\quoteda{a}\quoteda{n}\quoteda{d}\quoteda{ }\quoteda{t}\quoteda{h}\quoteda{e}\quoteda{ }\quoteda{A}\quoteda{m}\quoteda{e}\quoteda{r}\quoteda{i}\quoteda{c}\quoteda{a}\quoteda{s}\quoteda{,}\quotedb{ }\quotedb{c}\quoteda{a}\quoteda{t}\quoteda{e}\quoteda{r}\quoteda{i}\quoteda{n}\quoteda{g}\quoteda{ }\quoteda{t}\quoteda{o}\quoteda{ }\quoteda{t}\quoteda{h}\quoteda{e}\quoteda{ }\quoteda{t}\quoteda{a}\quoteda{s}\quoteda{t}\quoteda{e}\quoteda{s}\quoteda{ }\quoteda{o}\quoteda{f}\quoteda{ }chocolat\quoteda{e}\quoteda{ }\quoteda{l}\quoteda{o}\quoteda{v}\quoteda{e}\quoteda{r}\quoteda{s}\quoteda{ }\quoteda{a}\quoteda{r}\quoteda{o}\quoteda{u}\quoteda{n}\quoteda{d}\quoteda{ }\quoteda{t}\quoteda{h}\quoteda{e}\quoteda{ }\quoteda{w}\quoteda{o}\quoteda{r}\quoteda{l}\quoteda{d}\quoteda{.}
        & 44.7\\
        
        \cdashlinelr{1-2}
        \textbf{\method}: 
        Cadbury\quoteda{ }\quoteda{i}\quoteda{s}\quoteda{ }\quoteda{a}\quoteda{ }\quoteda{B}\quoteda{r}\quoteda{i}\quoteda{t}\quoteda{i}\quoteda{s}\quoteda{h}\quoteda{ }\quoteda{m}\quoteda{u}\quoteda{l}\quoteda{t}\quoteda{i}\quoteda{n}\quoteda{a}\quoteda{t}\quoteda{i}\quoteda{o}\quoteda{n}\quoteda{a}\quoteda{l}\quotedb{ }\quotedb{c}\quotedb{o}\quotedb{n}\quotedb{f}\quotedb{e}\quotedb{c}\quotedb{t}\quotedb{i}\quotedb{o}\quotedb{n}\quotedc{e}\quotedc{r}\quotedc{y}\quotedc{ }\quotedc{c}\quotedc{o}\quotedc{m}\quotedc{p}\quotedc{a}\quotedc{n}\quotedc{y}\quotedc{ }\quotedb{t}\quotedb{h}\quotedb{a}\quotedb{t}\quotedb{ }\quoteda{o}\quoteda{p}\quoteda{e}\quoteda{r}\quoteda{a}\quoteda{t}\quoteda{e}\quoteda{s}\quoteda{ }\quoteda{i}\quoteda{n}\quoteda{ }\quoteda{m}\quoteda{a}\quoteda{n}\quoteda{y}\quoteda{ }\quoteda{c}\quoteda{o}\quoteda{u}\quoteda{n}\quoteda{t}\quoteda{r}\quoteda{i}\quoteda{e}\quoteda{s}\quoteda{ }\quoteda{a}\quoteda{r}\quoteda{o}\quoteda{u}\quoteda{n}\quoteda{d}\quoteda{ }\quoteda{t}\quoteda{h}\quoteda{e}\quoteda{ }\quoteda{w}\quoteda{o}\quoteda{r}\quoteda{l}\quoteda{d}\quoteda{.}\quoteda{ }\quoteda{T}\quoteda{h}\quoteda{e}\quoteda{ }\quoteda{c}\quoteda{o}\quoteda{m}\quoteda{p}\quoteda{a}\quoteda{n}\quoteda{y}\quoteda{ }\quoteda{i}\quoteda{s}\quoteda{ }\quoteda{o}\quoteda{w}\quoteda{n}\quoteda{e}\quoteda{d}\quoteda{ }\quoteda{b}\quoteda{y}\quoteda{ }\quoteda{t}\quoteda{h}\quoteda{e}\quoteda{ }\quoteda{K}raft Foods\quoteda{ }\quoteda{G}\quoteda{r}\quoteda{o}\quoteda{u}\quoteda{p}\quotedb{ }\quotedb{a}\quotedb{n}\quotedb{d}\quotedb{ }\quotedb{h}\quotedb{a}\quotedb{s}\quotedb{ }\quotedb{a}\quotedb{ }\quotedb{s}\quotedb{i}\quotedb{g}\quotedb{n}\quotedb{i}\quotedb{f}\quotedb{i}\quotedb{c}\quotedb{a}\quotedb{n}\quotedb{t}\quotedb{ }\quoteda{p}\quoteda{r}\quoteda{e}\quoteda{s}\quoteda{e}\quoteda{n}\quoteda{c}\quoteda{e}\quoteda{ }\quoteda{i}\quoteda{n}\quotedb{ }\quotedb{m}\quotedb{a}\quotedb{r}\quotedb{k}\quotedb{e}\quotedb{t}\quoteda{s}\quoteda{ }\quoteda{i}\quoteda{n}\quoteda{ }\quoteda{E}\quoteda{u}\quoteda{r}\quoteda{o}\quoteda{p}\quoteda{e}\quoteda{,}\quoteda{ }\quoteda{N}\quoteda{o}\quoteda{r}\quoteda{t}\quoteda{h}\quoteda{ }\quoteda{A}\quoteda{m}\quoteda{e}\quoteda{r}\quoteda{i}\quoteda{c}\quoteda{a}\quoteda{,}\quoteda{ }\quoteda{A}\quoteda{s}\quoteda{i}\quoteda{a}\quotedb{,}\quotedb{ }\quotedb{a}\quotedb{n}\quotedb{d}\quotedb{ }\quotedb{A}\quotedb{u}\quotedb{s}\quotedb{t}\quotedb{r}\quotedb{a}\quotedb{l}\quotedb{i}\quotedb{a}\quotedb{.}\quotedb{ }\quotedb{A}\quoteda{c}\quoteda{c}\quoteda{o}\quoteda{r}\quoteda{d}\quoteda{i}\quoteda{n}\quoteda{g}\quoteda{ }\quoteda{t}\quoteda{o}\quoteda{ }\quoteda{t}\quoteda{h}\quoteda{e}\quoteda{ }\quoteda{c}\quoteda{o}\quoteda{m}\quoteda{p}\quoteda{a}\quoteda{n}\quoteda{y}\quoteda{'}\quoteda{s}\quoteda{ }\quoteda{w}\quoteda{e}\quoteda{b}\quoteda{s}\quoteda{i}\quoteda{t}\quoteda{e}\quoteda{,}\quoteda{ }Cadbury\quoteda{ }\quoteda{s}\quoteda{e}\quoteda{l}\quoteda{l}\quoteda{s}\quoteda{ }\quoteda{i}\quoteda{t}\quoteda{s}\quoteda{ }\quoteda{p}\quoteda{r}\quoteda{o}\quoteda{d}\quoteda{u}\quoteda{c}\quoteda{t}\quoteda{s}\quoteda{ }\quoteda{i}\quoteda{n}\quoteda{ }\quoteda{o}\quoteda{v}\quoteda{e}\quoteda{r}\quoteda{ }\quoteda{5}\quoteda{0}\quoteda{ }\quoteda{c}\quoteda{o}\quoteda{u}\quoteda{n}\quoteda{t}\quoteda{r}\quoteda{i}\quoteda{e}\quoteda{s}\quoteda{ }\quoteda{w}\quoteda{o}\quoteda{r}\quoteda{l}\quoteda{d}\quoteda{w}\quoteda{i}\quoteda{d}\quoteda{e}\quoteda{,}\quoteda{ }\quoteda{i}\quoteda{n}\quoteda{c}\quoteda{l}\quoteda{u}\quoteda{d}\quoteda{i}\quoteda{n}\quoteda{g}\quoteda{ }\quoteda{t}\quoteda{h}\quoteda{e}\quoteda{ }\quoteda{U}\quoteda{n}\quoteda{i}\quoteda{t}\quoteda{e}\quoteda{d}\quoteda{ }\quoteda{K}\quoteda{i}\quoteda{n}\quoteda{g}\quoteda{d}\quoteda{o}\quoteda{m}\quoteda{,}\quoteda{ }\quoteda{t}\quoteda{h}\quoteda{e}\quoteda{ }\quoteda{U}\quoteda{n}\quoteda{i}\quoteda{t}\quoteda{e}\quoteda{d}\quoteda{ }\quoteda{S}\quoteda{t}\quoteda{a}\quoteda{t}\quoteda{e}\quoteda{s}\quoteda{,}\quoteda{ }\quoteda{C}\quoteda{a}\quoteda{n}\quoteda{a}\quoteda{d}\quoteda{a}\quoteda{,}\quoteda{ }\quoteda{A}\quoteda{u}\quoteda{s}\quoteda{t}\quoteda{r}\quoteda{a}\quoteda{l}\quoteda{i}\quoteda{a}\quoteda{,}\quoteda{ }\quoteda{a}\quoteda{n}\quoteda{d}\quoteda{ }\quoteda{N}\quoteda{e}\quoteda{w}\quoteda{ }\quoteda{Z}\quoteda{e}\quoteda{a}\quoteda{l}\quoteda{a}\quoteda{n}\quoteda{d}\quoteda{.}
        &73.8\\
        
        \bottomrule
    \end{tabular}
    \caption{Additional example outputs generated by \textsc{Llama2-7B-Chat} before and after \method\ on NQ. Quoted segments are \quoteda{highlighted}. \quotedb{Lighter highlighting} and \quotedc{lightest highlighting} indicates two or three overlapped quoted segments, respectively. } 
    
    \label{tab:example5}
\end{table*}

\begin{table*}[ht]
    \centering
    \small
    \begin{tabular}{p{400pt}c}
        \toprule
        \textbf{\textit{Question:}} Where did the saying monkey's uncle come from? & \textbf{QUIP}\\

        \midrule
        \textbf{Reference}: 
        \quoteda{T}\quoteda{h}\quoteda{e}\quoteda{ }\quoteda{t}\quoteda{e}\quoteda{r}\quoteda{m}\quoteda{ }\quoteda{m}\quoteda{o}\quoteda{n}\quoteda{k}\quoteda{e}\quoteda{y}\quoteda{'}\quoteda{s}\quoteda{ }\quoteda{u}\quoteda{n}\quoteda{c}\quoteda{l}\quoteda{e}\quoteda{,}\quoteda{ }\quoteda{m}\quoteda{o}\quoteda{s}\quoteda{t}\quoteda{ }\quoteda{n}\quoteda{o}\quoteda{t}\quoteda{a}\quoteda{b}\quoteda{l}\quoteda{y}\quoteda{ }\quoteda{s}\quoteda{e}\quoteda{e}\quoteda{n}\quoteda{ }\quoteda{i}\quoteda{n}\quoteda{ }\quoteda{t}\quoteda{h}\quoteda{e}\quoteda{ }\quoteda{i}\quoteda{d}\quoteda{i}\quoteda{o}\quoteda{m}\quoteda{ }\quoteda{"}\quoteda{(}\quoteda{w}\quoteda{e}\quoteda{l}\quoteda{l}\quoteda{)}\quoteda{ }\quoteda{I} \quoteda{'}\quoteda{l}\quoteda{l}\quoteda{ }\quoteda{b}\quoteda{e}\quoteda{ }\quoteda{a}\quoteda{ }\quoteda{m}\quoteda{o}\quoteda{n}\quoteda{k}\quoteda{e}\quoteda{y}\quoteda{'}\quoteda{s}\quoteda{ }\quoteda{u}\quoteda{n}\quoteda{c}\quoteda{l}\quoteda{e}\quoteda{"}\quoteda{,}\quoteda{ }\quoteda{i}\quoteda{s}\quoteda{ }\quoteda{u}\quoteda{s}\quoteda{e}\quoteda{d}\quoteda{ }\quoteda{t}\quoteda{o}\quoteda{ }\quoteda{e}\quoteda{x}\quoteda{p}\quoteda{r}\quoteda{e}\quoteda{s}\quoteda{s}\quoteda{ }\quoteda{c}\quoteda{o}\quoteda{m}\quoteda{p}\quoteda{l}\quoteda{e}\quoteda{t}\quoteda{e}\quoteda{ }\quoteda{s}\quoteda{u}\quoteda{r}\quoteda{p}\quoteda{r}\quoteda{i}\quoteda{s}\quoteda{e}\quoteda{,}\quoteda{ }\quoteda{a}\quoteda{m}\quoteda{a}\quoteda{z}\quoteda{e}\quoteda{m}\quoteda{e}\quoteda{n}\quoteda{t}\quoteda{ }\quoteda{o}\quoteda{r}\quoteda{ }\quoteda{d}\quoteda{i}\quoteda{s}\quoteda{b}\quoteda{e}\quoteda{l}\quoteda{i}\quoteda{e}\quoteda{f}\quoteda{.}\quoteda{ }\quoteda{I}\quoteda{t}\quoteda{ }\quoteda{c}\quoteda{a}\quoteda{n}\quoteda{ }\quoteda{a}\quoteda{l}\quoteda{s}\quoteda{o}\quoteda{ }\quoteda{b}\quoteda{e}\quoteda{ }\quoteda{u}\quoteda{s}\quoteda{e}\quoteda{d}\quoteda{ }\quoteda{t}\quoteda{o}\quoteda{ }\quoteda{a}\quoteda{c}\quoteda{k}\quoteda{n}\quoteda{o}\quoteda{w}\quoteda{l}\quoteda{e}\quoteda{d}\quoteda{g}\quoteda{e}\quoteda{ }\quoteda{t}\quoteda{h}\quoteda{e}\quoteda{ }\quoteda{i}\quoteda{m}\quoteda{p}\quoteda{o}\quoteda{s}\quoteda{s}\quoteda{i}\quoteda{b}\quoteda{i}\quoteda{l}\quoteda{i}\quoteda{t}\quoteda{y}\quoteda{ }\quoteda{o}\quoteda{f}\quoteda{ }\quoteda{a}\quoteda{ }\quoteda{s}\quoteda{i}\quoteda{t}\quoteda{u}\quoteda{a}\quoteda{t}\quoteda{i}\quoteda{o}\quoteda{n}\quoteda{,}\quoteda{ }\quoteda{i}\quoteda{n}\quoteda{ }\quoteda{t}\quoteda{h}\quoteda{e}\quoteda{ }\quoteda{s}\quoteda{a}\quoteda{m}\quoteda{e}\quoteda{ }\quoteda{w}\quoteda{a}\quoteda{y}\quoteda{ }\quoteda{t}\quoteda{h}\quoteda{a}\quoteda{t}\quoteda{ }\quoteda{"}\quoteda{p}\quoteda{i}\quoteda{g}\quoteda{s}\quoteda{ }\quoteda{m}\quoteda{i}\quoteda{g}\quoteda{h}\quoteda{t}\quoteda{ }\quoteda{f}\quoteda{l}\quoteda{y}\quoteda{"}\quoteda{ }\quoteda{i}\quoteda{s}\quoteda{ }\quoteda{u}\quoteda{s}\quoteda{e}\quoteda{d}\quoteda{.}\quoteda{ }\quoteda{A}\quoteda{n}\quoteda{ }\quoteda{e}\quoteda{x}\quoteda{a}\quoteda{m}\quoteda{p}\quoteda{l}\quoteda{e}\quoteda{ }\quoteda{i}\quoteda{s}\quoteda{ }\quoteda{i}\quoteda{f}\quoteda{ }\quoteda{o}\quoteda{n}\quoteda{e}\quoteda{ }\quoteda{s}\quoteda{a}\quoteda{y}\quoteda{s}\quoteda{:}\quoteda{ }\quoteda{"}\quoteda{I}\quoteda{ }\quoteda{m}\quoteda{a}\quoteda{y}\quoteda{ }\quoteda{a}\quoteda{g}\quoteda{r}\quoteda{e}\quoteda{e}\quoteda{ }\quoteda{t}\quoteda{h}\quoteda{a}\quoteda{t}\quoteda{ }\quoteda{i}\quoteda{f}\quoteda{ }\quoteda{t}\quoteda{w}\quoteda{o}\quoteda{ }\quoteda{p}\quoteda{l}\quoteda{u}\quoteda{s}\quoteda{ }\quoteda{t}\quoteda{w}\quoteda{o}\quoteda{ }\quoteda{e}\quoteda{q}\quoteda{u}\quoteda{a}\quoteda{l}\quoteda{s}\quoteda{ }\quoteda{f}\quoteda{i}\quoteda{v}\quoteda{e}\quoteda{,}\quoteda{ }\quoteda{t}\quoteda{h}\quoteda{e}\quoteda{n}\quoteda{ }\quoteda{I}\quoteda{ }\quoteda{a}\quoteda{m}\quoteda{ }\quoteda{a}\quoteda{ }\quoteda{m}\quoteda{o}\quoteda{n}\quoteda{k}\quoteda{e}\quoteda{y}\quoteda{'}\quoteda{s}\quoteda{ }\quoteda{u}\quoteda{n}\quoteda{c}\quoteda{l}\quoteda{e}\quoteda{"}\quoteda{.} "I 'll be a monkey's uncle"\quoteda{ }\quoteda{h}\quoteda{a}\quoteda{s}\quoteda{ }\quoteda{b}\quoteda{e}\quoteda{e}\quoteda{n}\quoteda{ }\quoteda{s}\quoteda{a}\quoteda{i}\quoteda{d}\quoteda{ }\quoteda{t}\quoteda{o}\quoteda{ }\quoteda{d}\quoteda{a}\quoteda{t}\quoteda{e}\quoteda{ }\quoteda{f}\quoteda{r}\quoteda{o}\quoteda{m}\quoteda{ }\quoteda{a}fter 1925, the dat\quoteda{e}\quoteda{ }\quoteda{o}\quoteda{f}\quoteda{ }\quoteda{t}\quoteda{h}\quoteda{e}\quoteda{ }\quoteda{w}\quoteda{i}\quoteda{d}\quoteda{e}\quoteda{l}\quoteda{y}\quoteda{ }\quoteda{p}\quoteda{u}\quoteda{b}\quoteda{l}\quoteda{i}\quoteda{c}\quoteda{i}\quoteda{z}\quoteda{e}\quoteda{d}\quoteda{ }\quoteda{S}\quoteda{c}\quoteda{o}\quoteda{p}\quoteda{e}\quoteda{s}\quoteda{ }\quoteda{T}\quoteda{r}\quoteda{i}\quoteda{a}\quoteda{l}\quoteda{ }\quoteda{i}\quoteda{n}\quoteda{ }\quoteda{t}\quoteda{h}\quoteda{e}\quoteda{ }\quoteda{U}\quoteda{n}\quoteda{i}\quoteda{t}\quoteda{e}\quoteda{d}\quoteda{ }\quoteda{S}\quoteda{t}\quoteda{a}\quoteda{t}\quoteda{e}\quoteda{s}\quoteda{,}\quoteda{ }\quoteda{w}\quoteda{h}\quoteda{e}\quoteda{r}\quoteda{e}\quoteda{ }\quoteda{t}\quoteda{h}\quoteda{e}\quoteda{ }\quoteda{t}\quoteda{e}\quoteda{r}\quoteda{m}\quoteda{ }\quoteda{f}\quoteda{i}\quoteda{r}\quoteda{s}\quoteda{t}\quoteda{ }\quoteda{a}\quoteda{p}\quoteda{p}\quoteda{e}\quotedb{a}\quotedb{r}\quotedb{s}\quotedb{.}\quotedb{ }\quotedb{T}\quotedb{h}\quotedb{e}\quotedb{ }\quoteda{O}\quoteda{x}\quoteda{f}\quoteda{o}\quoteda{r}\quoteda{d}\quoteda{ }\quoteda{E}\quoteda{n}\quoteda{g}\quoteda{l}\quoteda{i}\quoteda{s}\quoteda{h}\quoteda{ }\quoteda{D}\quoteda{i}\quoteda{c}\quoteda{t}\quoteda{i}\quoteda{o}\quoteda{n}\quoteda{a}\quoteda{r}\quoteda{y}\quoteda{'}\quoteda{s}\quotedb{ }\quotedb{e}\quotedb{a}\quotedb{r}\quotedb{l}\quotedb{i}\quotedb{e}\quotedb{s}\quotedb{t}\quotedb{ }\quoteda{e}\quoteda{x}\quoteda{a}\quoteda{m}\quoteda{p}\quoteda{l}\quoteda{e}\quoteda{ }\quoteda{i}\quoteda{s}\quoteda{ }\quoteda{t}\quoteda{h}\quoteda{e}\quoteda{ }\quoteda{p}hrase If that's a joke I 'm a monkey's uncle\quoteda{,}\quoteda{ }\quoteda{f}\quoteda{r}\quoteda{o}\quoteda{m}\quoteda{ }\quoteda{a}\quoteda{n}\quoteda{ }\quoteda{O}\quoteda{h}\quoteda{i}\quoteda{o}\quoteda{ }\quoteda{n}\quoteda{e}\quoteda{w}\quoteda{s}\quoteda{p}\quoteda{a}\quoteda{p}\quoteda{e}\quoteda{r}\quoteda{ }\quoteda{o}\quoteda{n}\quoteda{ }\quoteda{8}\quoteda{ }\quoteda{F}\quoteda{e}\quoteda{b}\quoteda{r}\quoteda{u}\quoteda{a}\quoteda{r}\quoteda{y}\quoteda{ }\quoteda{1}\quoteda{9}\quoteda{2}\quoteda{5}\quoteda{.}\quoteda{ }\quoteda{I}\quoteda{t}\quoteda{ }\quoteda{w}\quoteda{a}\quoteda{s}\quoteda{ }\quoteda{o}\quoteda{r}\quoteda{i}\quoteda{g}\quoteda{i}\quoteda{n}\quoteda{a}\quoteda{l}\quoteda{l}\quoteda{y}\quoteda{ }\quoteda{a}\quoteda{ }\quoteda{s}\quoteda{a}\quoteda{r}\quoteda{c}\quoteda{a}\quoteda{s}\quoteda{t}\quoteda{i}\quoteda{c}\quoteda{ }\quoteda{r}\quoteda{e}\quoteda{m}\quoteda{a}\quoteda{r}\quoteda{k}\quoteda{ }\quoteda{m}\quoteda{a}\quoteda{d}\quoteda{e}\quoteda{ }\quoteda{b}\quoteda{y}\quoteda{ }\quoteda{c}\quoteda{r}\quoteda{e}\quoteda{a}\quoteda{t}\quoteda{i}\quoteda{o}\quoteda{n}\quoteda{i}\quoteda{s}\quoteda{t}\quoteda{s}\quoteda{.}\quoteda{ }\quoteda{T}\quoteda{h}\quoteda{e}\quoteda{ }\quoteda{n}\quoteda{o}\quoteda{t}\quoteda{i}\quoteda{o}\quoteda{n}\quoteda{ }\quoteda{"}\quoteda{t}\quoteda{h}\quoteda{a}\quoteda{t}\quoteda{ }(people)\quoteda{ }\quoteda{w}\quoteda{e}\quoteda{r}\quoteda{e}\quoteda{ }\quoteda{d}\quoteda{e}\quoteda{s}\quoteda{c}\quoteda{e}\quoteda{n}\quoteda{d}\quoteda{e}\quoteda{d}\quoteda{ }\quoteda{f}\quoteda{r}\quoteda{o}\quoteda{m}\quoteda{ }\quoteda{a}\quoteda{p}\quoteda{e}\quoteda{s}\quoteda{ }\quoteda{w}\quoteda{a}\quoteda{s}\quoteda{ }\quoteda{c}\quoteda{o}\quoteda{n}\quoteda{s}\quoteda{i}\quoteda{d}\quoteda{e}\quoteda{r}\quoteda{e}\quoteda{d}\quoteda{ }\quoteda{b}\quoteda{l}\quoteda{a}\quoteda{s}\quoteda{p}\quoteda{h}\quoteda{e}\quoteda{m}\quoteda{o}\quoteda{u}\quoteda{s}\quoteda{.}\quoteda{.}\quoteda{.}\quoteda{ }\quoteda{b}\quoteda{y}\quoteda{ }\quoteda{D}\quoteda{a}\quoteda{r}\quoteda{w}\quoteda{i}\quoteda{n}\quoteda{'}\quoteda{s}\quoteda{ }\quoteda{c}\quoteda{o}\quoteda{n}\quoteda{t}\quoteda{e}\quoteda{m}\quoteda{p}\quoteda{o}\quoteda{r}\quoteda{a}\quoteda{r}\quoteda{i}\quoteda{e}\quoteda{s}\quoteda{"}\quoteda{,}\quoteda{ }\quoteda{a}\quoteda{n}\quoteda{d}\quoteda{ }\quoteda{i}\quoteda{t}\quoteda{ }\quoteda{w}\quoteda{a}\quoteda{s}\quoteda{ }\quoteda{f}\quoteda{o}\quoteda{r}\quoteda{ }\quoteda{t}\quoteda{h}\quoteda{i}\quoteda{s}\quoteda{ }\quoteda{r}\quoteda{e}\quoteda{a}\quoteda{s}\quoteda{o}\quoteda{n}\quoteda{ }\quoteda{t}\quoteda{h}\quoteda{a}\quoteda{t}\quoteda{ }\quoteda{t}\quoteda{h}\quoteda{e}\quoteda{ }\quoteda{s}\quoteda{a}\quoteda{r}\quoteda{c}\quoteda{a}\quoteda{s}\quoteda{t}\quoteda{i}\quoteda{c}\quoteda{ }\quoteda{p}\quoteda{h}\quoteda{r}\quoteda{a}\quoteda{s}\quoteda{e}\quoteda{ }\quoteda{c}\quoteda{a}\quoteda{m}\quoteda{e}\quoteda{ }\quoteda{i}\quoteda{n}\quoteda{t}\quoteda{o}\quoteda{ }\quoteda{u}\quoteda{s}\quoteda{e}\quoteda{.}
        & 69.6\\
        
        \cdashlinelr{1-2}
        \textsc{\textbf{Llama2-7B-Chat}}: 
        \quoteda{T}\quoteda{h}\quoteda{e}\quoteda{ }\quoteda{o}\quoteda{r}\quoteda{i}\quoteda{g}\quoteda{i}\quoteda{n}\quoteda{s}\quoteda{ }\quoteda{o}\quoteda{f}\quoteda{ }\quoteda{t}\quoteda{h}\quoteda{e}\quoteda{ }\quoteda{p}\quoteda{h}\quoteda{r}\quoteda{a}\quoteda{s}\quoteda{e}\quoteda{ }\quoteda{"}monkey's uncle"\quoteda{ }\quoteda{a}\quoteda{r}\quoteda{e}\quoteda{ }\quoteda{u}\quoteda{n}\quoteda{c}\quoteda{e}\quoteda{r}\quoteda{t}\quoteda{a}\quoteda{i}\quoteda{n}\quoteda{,}\quoteda{ }\quoteda{b}\quoteda{u}\quoteda{t}\quoteda{ }\quoteda{i}\quoteda{t}\quoteda{ }\quoteda{i}\quoteda{s}\quoteda{ }\quoteda{b}\quoteda{e}\quoteda{l}\quoteda{i}\quoteda{e}\quoteda{v}\quoteda{e}\quoteda{d}\quoteda{ }\quoteda{t}\quoteda{o}\quoteda{ }\quoteda{h}\quoteda{a}\quoteda{v}\quoteda{e}\quoteda{ }\quoteda{o}\quoteda{r}\quoteda{i}\quoteda{g}\quoteda{i}\quoteda{n}\quoteda{a}\quoteda{t}\quoteda{e}\quoteda{d}\quoteda{ }\quoteda{i}\quoteda{n}\quoteda{ }\quoteda{t}\quoteda{h}\quoteda{e}\quoteda{ }\quoteda{U}\quoteda{n}\quoteda{i}\quoteda{t}\quoteda{e}\quoteda{d}\quoteda{ }\quoteda{S}\quoteda{t}\quoteda{a}\quoteda{t}\quoteda{e}\quoteda{s}\quoteda{ }\quoteda{i}\quoteda{n}\quoteda{ }\quoteda{t}\quoteda{h}\quoteda{e}\quoteda{ }\quoteda{m}\quoteda{i}\quoteda{d}\quoteda{-}\quoteda{t}\quoteda{o}\quoteda{-}\quoteda{l}\quoteda{a}\quoteda{t}\quotedb{e}\quotedb{ }\quotedb{1}\quotedb{9}\quotedb{t}\quotedb{h}\quotedb{ }\quotedb{c}\quotedb{e}\quotedb{n}\quotedb{t}\quotedb{u}\quotedb{r}\quotedb{y}\quotedb{.}\quotedb{ }\quotedb{O}\quotedb{n}\quotedb{e}\quotedb{ }\quoteda{t}\quoteda{h}\quoteda{e}\quoteda{o}\quoteda{r}\quotedb{y}\quotedb{ }\quotedb{i}\quotedb{s}\quotedb{ }\quotedb{t}\quotedb{h}\quotedb{a}\quotedb{t}\quotedb{ }\quotedb{i}\quotedc{t}\quotedc{ }\quotedc{w}\quotedc{a}\quotedc{s}\quotedc{ }\quotedc{u}\quotedc{s}\quotedc{e}\quotedc{d}\quotedc{ }\quotedb{a}\quotedb{s}\quotedb{ }\quotedb{a}\quotedb{ }\quotedb{w}\quotedb{a}\quotedb{y}\quotedb{ }\quotedb{t}\quotedb{o}\quotedb{ }\quoteda{e}\quoteda{x}\quoteda{p}\quoteda{r}\quoteda{e}\quoteda{s}\quoteda{s}\quoteda{ }\quoteda{d}\quoteda{i}\quoteda{s}belief or skepticism\quoteda{,}\quoteda{ }\quoteda{s}\quoteda{i}\quoteda{m}\quoteda{i}\quoteda{l}\quoteda{a}\quoteda{r}\quoteda{ }\quoteda{t}\quoteda{o}\quoteda{ }\quoteda{t}\quoteda{h}\quoteda{e}\quoteda{ }\quoteda{p}\quoteda{h}\quoteda{r}\quoteda{a}\quoteda{s}\quoteda{e}\quoteda{ }\quoteda{"}\quoteda{c}lose but no cigar\quoteda{.}\quoteda{"}\quoteda{ }\quoteda{A}\quoteda{n}\quoteda{o}\quoteda{t}\quoteda{h}\quoteda{e}\quoteda{r}\quoteda{ }\quoteda{t}\quoteda{h}\quoteda{e}\quoteda{o}\quotedb{r}\quotedb{y}\quotedb{ }\quotedb{s}\quotedb{u}\quotedb{g}\quotedb{g}\quotedb{e}\quotedb{s}\quotedb{t}\quotedb{s}\quotedb{ }\quotedb{t}\quotedb{h}\quotedb{a}\quotedb{t}\quotedb{ }\quotedb{i}\quotedb{t}\quotedc{ }\quotedc{m}\quotedb{a}\quotedb{y}\quotedb{ }\quotedb{h}\quotedb{a}\quotedb{v}\quotedb{e}\quotedb{ }\quotedb{b}\quotedb{e}\quotedb{e}\quotedb{n}\quotedb{ }\quotedb{a}\quotedb{ }\quotedb{t}\quotedb{e}\quoteda{r}\quoteda{m}\quoteda{ }\quoteda{u}\quoteda{s}\quoteda{e}\quoteda{d} in vaudeville or minstrel shows\quoteda{ }\quoteda{t}\quoteda{o}\quoteda{ }\quoteda{r}\quoteda{e}\quoteda{f}\quoteda{e}\quoteda{r}\quoteda{ }\quoteda{t}\quoteda{o}\quoteda{ }\quoteda{a}\quoteda{ }\quoteda{b}\quoteda{l}\quoteda{a}\quoteda{c}\quoteda{k}\quoteda{ }\quoteda{p}\quoteda{e}\quoteda{r}\quoteda{s}\quoteda{o}\quoteda{n}\quoteda{,}\quoteda{ }with\quoteda{ }\quoteda{t}\quoteda{h}\quoteda{e}\quoteda{ }\quoteda{p}\quoteda{h}\quoteda{r}\quoteda{a}\quoteda{s}\quoteda{e}\quoteda{ }\quoteda{b}\quoteda{e}\quoteda{i}\quoteda{n}\quoteda{g}\quoteda{ }\quoteda{u}\quoteda{s}\quoteda{e}\quoteda{d}\quoteda{ }\quoteda{a}\quoteda{s}\quoteda{ }\quoteda{a}\quoteda{ }mild insult or racial slu\quoteda{r}\quoteda{.}\quoteda{ }\quoteda{R}\quoteda{e}\quoteda{g}\quoteda{a}\quoteda{r}\quoteda{d}\quoteda{l}\quoteda{e}\quoteda{s}\quoteda{s}\quoteda{ }\quoteda{o}\quoteda{f}\quoteda{ }\quoteda{i}\quoteda{t}\quoteda{s}\quoteda{ }\quoteda{o}\quoteda{r}\quoteda{i}\quoteda{g}\quoteda{i}\quoteda{n}\quoteda{s}\quoteda{,}\quoteda{ }"monkey's uncle"\quoteda{ }\quoteda{h}\quoteda{a}\quoteda{s}\quoteda{ }\quoteda{b}\quoteda{e}\quoteda{c}\quoteda{o}\quoteda{m}\quoteda{e}\quoteda{ }\quoteda{a}\quoteda{ }\quoteda{c}\quoteda{o}\quoteda{m}\quoteda{m}\quoteda{o}\quoteda{n}\quoteda{ }\quoteda{a}\quoteda{n}\quoteda{d}\quoteda{ }\quoteda{e}nduring idiom used to express\quoteda{ }\quoteda{i}\quoteda{n}\quoteda{c}\quoteda{r}\quoteda{e}\quoteda{d}\quoteda{u}\quoteda{l}\quoteda{i}\quoteda{t}\quoteda{y}\quoteda{ }\quoteda{o}\quoteda{r}\quoteda{ }\quoteda{d}\quoteda{i}\quoteda{s}\quoteda{b}\quoteda{e}\quoteda{l}\quoteda{i}\quoteda{e}\quoteda{f}\quoteda{.}
        & 33.0\\
        
        \cdashlinelr{1-2}
        \textbf{\method}: 
        The origin of\quoteda{ }\quoteda{t}\quoteda{h}\quoteda{e}\quoteda{ }\quoteda{i}\quoteda{d}\quoteda{i}\quoteda{o}\quoteda{m}\quoteda{ }\quoteda{"}\quoteda{m}\quoteda{o}\quoteda{n}\quoteda{k}\quoteda{e}\quoteda{y}\quoteda{'}\quoteda{s}\quoteda{ }\quoteda{u}\quoteda{n}\quoteda{c}\quoteda{l}\quoteda{e}\quoteda{"}\quotedb{ }\quoteda{i}\quoteda{s}\quoteda{ }\quoteda{u}\quoteda{n}\quoteda{c}\quoteda{e}\quoteda{r}\quoteda{t}\quoteda{a}\quoteda{i}\quoteda{n}\quoteda{,}\quoteda{ }\quoteda{b}\quoteda{u}\quoteda{t}\quoteda{ }\quoteda{i}\quoteda{t}\quoteda{ }\quoteda{i}\quoteda{s}\quoteda{ }\quoteda{b}\quoteda{e}\quoteda{l}\quoteda{i}\quoteda{e}\quoteda{v}\quoteda{e}\quoteda{d}\quoteda{ }\quoteda{t}\quoteda{o}\quoteda{ }\quoteda{h}\quoteda{a}\quoteda{v}\quoteda{e}\quoteda{ }\quoteda{o}\quoteda{r}\quoteda{i}\quoteda{g}\quoteda{i}\quoteda{n}\quoteda{a}\quoteda{t}\quoteda{e}\quoteda{d}\quoteda{ }\quoteda{i}\quoteda{n}\quoteda{ }\quoteda{t}\quoteda{h}\quoteda{e}\quoteda{ }\quoteda{U}\quoteda{n}\quoteda{i}\quoteda{t}\quoteda{e}\quoteda{d}\quoteda{ }\quoteda{S}\quoteda{t}\quoteda{a}\quoteda{t}\quoteda{e}\quoteda{s}\quoteda{ }\quoteda{i}\quoteda{n}\quoteda{ }\quoteda{t}\quoteda{h}\quoteda{e}\quoteda{ }\quoteda{l}\quoteda{a}\quoteda{t}\quoteda{e}\quoteda{ }\quoteda{1}\quoteda{9}\quoteda{t}\quoteda{h}\quoteda{ }\quoteda{o}\quoteda{r}\quoteda{ }\quoteda{e}\quoteda{a}\quoteda{r}\quoteda{l}\quoteda{y}\quoteda{ }\quoteda{2}\quoteda{0}\quoteda{t}\quoteda{h}\quoteda{ }\quoteda{c}\quoteda{e}\quoteda{n}\quoteda{t}\quoteda{u}\quoteda{r}\quotedb{y}\quotedb{.}\quotedb{ }\quotedc{T}\quotedc{h}\quotedc{e}\quotedc{ }\quotedc{e}\quotedc{x}\quotedb{a}\quotedb{c}\quotedb{t}\quotedb{ }\quotedb{m}\quotedb{e}\quotedb{a}\quotedb{n}\quotedb{i}\quotedb{n}\quotedb{g}\quotedb{ }\quotedb{a}\quotedb{n}\quotedb{d}\quotedb{ }\quotedb{u}\quoteda{s}\quoteda{a}\quoteda{g}\quoteda{e}\quotedb{ }\quotedb{o}\quotedb{f}\quotedb{ }\quotedb{t}\quotedb{h}\quotedb{e}\quotedb{ }\quoteda{p}\quoteda{h}\quoteda{r}\quoteda{a}\quoteda{s}\quoteda{e}\quoteda{ }\quoteda{a}\quoteda{r}\quoteda{e}\quoteda{ }\quoteda{a}\quoteda{l}\quoteda{s}\quoteda{o}\quotedb{ }\quotedb{u}\quotedb{n}\quoteda{c}\quoteda{l}\quoteda{e}\quoteda{a}\quoteda{r}\quoteda{,}\quoteda{ }\quoteda{b}\quoteda{u}\quoteda{t}\quotedb{ }\quotedb{i}\quotedb{t}\quotedb{ }\quotedb{i}\quotedb{s}\quotedb{ }\quotedb{t}\quotedb{h}\quotedb{o}\quotedb{u}\quotedb{g}\quotedb{h}\quotedb{t}\quotedb{ }\quotedb{t}\quotedb{o}\quotedb{ }\quotedb{b}\quotedc{e}\quotedc{ }\quotedc{u}\quotedb{s}\quotedb{e}\quotedb{d}\quotedb{ }\quotedb{a}\quotedb{s}\quotedb{ }\quotedb{a}\quotedb{ }\quotedb{w}\quoteda{a}\quoteda{y}\quoteda{ }\quoteda{o}\quoteda{f}\quoteda{ }\quoteda{e}\quoteda{x}\quoteda{p}\quoteda{r}\quoteda{e}\quoteda{s}\quoteda{s}\quoteda{i}\quoteda{n}\quoteda{g}\quoteda{ }\quoteda{s}urprise or disbelief,\quoteda{ }\quoteda{s}\quoteda{i}\quoteda{m}\quoteda{i}\quoteda{l}\quoteda{a}\quoteda{r}\quoteda{ }\quoteda{t}\quoteda{o}\quoteda{ }\quoteda{o}\quoteda{t}\quoteda{h}\quoteda{e}\quoteda{r}\quoteda{ }\quoteda{i}\quoteda{d}\quoteda{i}\quoteda{o}\quoteda{m}\quoteda{s}\quoteda{ }\quoteda{s}\quoteda{u}\quoteda{c}\quoteda{h}\quoteda{ }\quoteda{a}\quoteda{s}\quoteda{ }"oh brother" or "good grief"\quoteda{.}\quoteda{ }\quoteda{T}\quoteda{h}\quoteda{e}\quoteda{ }\quoteda{p}\quoteda{h}\quoteda{r}\quoteda{a}\quoteda{s}\quoteda{e}\quoteda{ }\quoteda{h}\quoteda{a}\quoteda{s}\quoteda{ }\quoteda{s}\quoteda{i}\quoteda{n}\quoteda{c}\quoteda{e}\quoteda{ }\quoteda{b}\quoteda{e}\quoteda{c}\quoteda{o}\quotedb{m}\quotedb{e}\quotedb{ }\quotedb{c}\quotedb{o}\quotedb{m}\quotedb{m}\quotedb{o}\quotedb{n}\quotedc{l}\quotedc{y}\quotedc{ }\quotedc{u}\quotedc{s}\quotedc{e}\quotedc{d}\quotedc{ }\quotedc{i}\quotedc{n}\quotedc{ }\quotedb{v}\quotedb{a}\quotedb{r}\quotedb{i}\quotedb{o}\quotedb{u}\quotedb{s}\quotedb{ }\quotedb{p}\quoteda{a}\quoteda{r}\quoteda{t}\quoteda{s}\quoteda{ }\quoteda{o}\quoteda{f}\quoteda{ }\quoteda{t}\quoteda{h}\quoteda{e}\quoteda{ }\quoteda{E}\quoteda{n}\quoteda{g}\quoteda{l}\quoteda{i}\quoteda{s}\quoteda{h}\quoteda{-}\quoteda{s}\quoteda{p}\quoteda{e}\quoteda{a}\quoteda{k}\quoteda{i}\quoteda{n}\quoteda{g}\quotedb{ }\quotedb{w}\quotedb{o}\quotedb{r}\quotedb{l}\quotedb{d}\quotedc{ }\quotedc{a}\quotedc{n}\quotedc{d}\quotedc{ }\quotedc{h}\quotedc{a}\quotedc{s}\quotedc{ }\quotedc{b}\quotedc{e}\quotedc{e}\quotedc{n}\quotedc{ }\quotedb{u}\quotedb{s}\quotedb{e}\quotedb{d}\quotedb{ }\quotedb{i}\quoteda{n}\quoteda{ }\quoteda{p}\quoteda{o}\quoteda{p}\quoteda{u}\quoteda{l}\quoteda{a}\quoteda{r}\quoteda{ }\quoteda{c}\quoteda{u}\quoteda{l}\quoteda{t}\quoteda{u}\quoteda{r}\quoteda{e}\quotedb{,}\quotedb{ }\quotedb{i}\quotedb{n}\quotedb{c}\quotedb{l}\quotedb{u}\quotedb{d}\quotedc{i}\quotedc{n}\quotedc{g}\quotedc{ }\quotedc{i}\quotedc{n}\quotedc{ }\quotedc{l}\quotedc{i}\quotedc{t}\quotedc{e}\quotedc{r}\quotedc{a}\quotedc{t}\quotedc{u}\quotedc{r}\quotedc{e}\quotedc{,}\quotedc{ }\quotedb{m}\quotedb{u}\quotedb{s}\quotedb{i}\quotedb{c}\quotedb{,}\quotedb{ }\quoteda{a}\quoteda{n}\quoteda{d}\quoteda{ }\quoteda{f}\quoteda{i}\quoteda{l}\quoteda{m}\quoteda{.}
        &52.9\\
        
        \bottomrule
    \end{tabular}
    \caption{Additional example outputs generated by \textsc{Llama2-7B-Chat} before and after \method\ on NQ. Quoted segments are \quoteda{highlighted}. \quotedb{Lighter highlighting} and \quotedc{lightest highlighting} indicates two or three overlapped quoted segments, respectively. } 
    
    \label{tab:example3}
\end{table*}

\end{document}